\documentclass{article}

\PassOptionsToPackage{numbers, compress}{natbib}

\usepackage[preprint]{NIPS2025_Arxiv/cpal_2025}

\usepackage[utf8]{inputenc} 
\usepackage[T1]{fontenc}    
\usepackage[hidelinks,colorlinks=true,linkcolor=blue,citecolor=blue]{hyperref}       
\usepackage{url}            
\usepackage{xurl}   
\usepackage{booktabs}       
\usepackage{amsfonts}       
\usepackage{nicefrac}       
\usepackage{microtype}      

\usepackage[table]{xcolor}
\usepackage{array}      

\hypersetup{
    linkcolor=blue,
    citecolor=blue,      
    urlcolor=blue,
    pdfpagemode=FullScreen,
}

\usepackage{url}

\usepackage{amsmath,amsthm,amssymb,amsbsy,amsfonts,amscd,bm}
\usepackage{paralist}
\usepackage{xcolor}
\usepackage{color}
\usepackage{graphicx}
\graphicspath{{./figs/}}
\usepackage{comment}
\usepackage{multirow}
\usepackage{enumitem}
\usepackage{fancyhdr}
\usepackage{subcaption}
\usepackage{wrapfig}

\newtheorem{assum}{Assumption}

\newtheorem{theorem}{Theorem}

\newcommand{\e}{\begin{equation}}
\newcommand{\ee}{\end{equation}}
\newcommand{\en}{\begin{equation*}}
\newcommand{\een}{\end{equation*}}
\newcommand{\eqn}{\begin{eqnarray}}
\newcommand{\eeqn}{\end{eqnarray}}
\newcommand{\bmat}{\begin{bmatrix}}
\newcommand{\emat}{\end{bmatrix}}

\DeclareMathAlphabet\mathbfcal{OMS}{cmsy}{b}{n}

\newcommand{\mb}{\bm}

\newcommand{\diag}{\operatorname{diag}}

\DeclareMathOperator*{\argmin}{\text{arg~min}}
\DeclareMathOperator*{\argmax}{\text{arg~max}}

\newcommand{ \paren }[1]{ \left( #1 \right) }

\graphicspath{{./figs/}}

\newlength{\imgwidth}
\newboolean{twoColVersion}
\setboolean{twoColVersion}{false}
\newcommand{\twoCol}[2]{\ifthenelse{\boolean{twoColVersion}} {#1} {#2} }

\usepackage{cleveref}

\title{Towards Understanding the Mechanisms of Classifier-Free Guidance}

\author{%
   Xiang Li\textsuperscript{1} ~Rongrong Wang\textsuperscript{2} ~Qing Qu
   \textsuperscript{1} \\
  \textsuperscript{1}Department of EECS, University of Michigan \\
  \textsuperscript{2}Department of CMSE and Mathematics, Michigan State University \\
  \texttt{forkobe@umich.edu, wangron6@msu.edu, qingqu@umich.edu}
}

\begin{document}

\maketitle

\begin{abstract}
Classifier-free guidance (CFG) is a core technique powering state-of-the-art image generation systems, yet its underlying mechanisms remain poorly understood. In this work, we begin by analyzing CFG in a simplified linear diffusion model, where we show its behavior closely resembles that observed in the nonlinear case. Our analysis reveals that linear CFG improves generation quality via three distinct components: (i) a mean-shift term that approximately steers samples in the direction of class means, (ii) a positive Contrastive Principal Components (CPC) term that amplifies class-specific features, and (iii) a negative CPC term that suppresses generic features prevalent in unconditional data. We then verify these insights in real-world, \emph{nonlinear} diffusion models: over a broad range of noise levels, linear CFG resembles the behavior of its nonlinear counterpart. Although the two eventually diverge at low noise levels, we discuss how the insights from the linear analysis still shed light on the CFG's mechanism in the nonlinear regime.
\end{abstract}

\tableofcontents

\section{Introduction}
Diffusion models~\cite{ho2020denoising, songdenoising, songscore, karras2022elucidating} generate samples from a data distribution $p_\text{data}(\mb x)$, where $\mb x\in\mathbb{R}^d$, by reversing a forward noising process. This forward process, defined in~\eqref{forward process}, progressively corrupts the clean data until $p(\mb x;\sigma_\text{max})$ becomes indistinguishable from a Gaussian distribution $\mathcal{N}(\mb 0,\sigma_\text{max}^2\mb I)$,
\begin{align}
    \label{forward process}
    p(\mb x;\sigma(t))=\int_{\mathbb{R}^d}p_{0t}(\mb x|\mb x_0)p_\text{data}(\mb x_0)d\mb x_0.
\end{align}
Following the state-of-the-art EDM framework~\cite{karras2022elucidating,karras2024analyzing},  the forward transition kernel is set to $p_{0t}(\mb x|\mb x_0)=\mathcal{N}(\mb x_0,\sigma^2(t)\mb I)$. The reverse process can then be expressed as a probabilistic ODE:
\begin{align}
    \label{reverse ode}
    d\mb x_t = -\sigma(t)\nabla_{\mb x_t}\log p(\mb x;\sigma(t))dt,
\end{align}
such that $\mb x_t\sim p(\mb x;\sigma(t))$ for every $\sigma(t)\in (0,\sigma_\text{max}]$. In practice, the score function can be approximated as $\nabla_{\mb x}\log p(\mb x;\sigma(t))\approx (\mathcal{D_{\mb \theta}}(\mb x;\sigma(t))-\mb x)/\sigma^2(t)$, where $\mathcal{\mb D}_{\mb \theta}$ is a deep network-based denoiser with parameter $\mb \theta$ optimized by minimizing the denoising score matching objective~\cite{6795935}:
\begin{align}
    \label{denoising score matching}
    \mathbb{E}_{\mb x\sim p_\text{data},\mb\epsilon\sim\mathcal{N}(\mb 0,\sigma^2(t)\mb I)}[||\mathcal{D}_{\mb \theta}(\mb x+\mb \epsilon;\sigma(t))-\mb x||_2^2].
\end{align}
To sample from conditional distribution $p(\mb x|\mb c)$, the deep denoiser $\mathcal{D}_{\mb\theta}(\mb x;\sigma(t),\mb c)$ receives an auxiliary embedding $\mb c$ specifying the target class or other conditions during training such that conditional sampling can be performed with:
\begin{align}
    \label{conditional sampling}
    d\mb x_t = -\sigma(t)\nabla_{\mb x_t}\log p(\mb x|
    \mb c;\sigma(t))dt,
\end{align}
where $\nabla_{\mb x}\log p(\mb x|\mb c;\sigma(t))\approx (\mathcal{D_{\mb \theta}}(\mb x;\sigma(t),\mb c)-\mb x)/\sigma^2(t)$. 
However, the naive (standard) conditional sampling~\eqref{conditional sampling} alone often results in images with incoherent structures and fail to align well with the target condition~\cite{bradley2024classifier}. Classifier-free guidance (CFG)~\cite{ho2021classifier} addresses this issue by steering the naive conditional sampling trajectory with a \emph{guidance term}:
\begin{align}
    g(\mb x, t) = \nabla_{\mb x}\log p(\mb x|\mb c;\sigma(t))-\nabla_{\mb x}\log p(\mb x;\sigma(t)),
    \label{cfg guidance term}
\end{align}
so that~\eqref{conditional sampling} becomes:
\begin{align}\label{cfg sampling}
    d\mb x_t = -\sigma(t)(\nabla_{\mb x_t}\log p(\mb x|
    \mb c;\sigma(t))+\gamma  g(\mb x, t))dt,
\end{align}

where $\gamma\geq 0$ controls the strength of guidance. With a properly chosen $\gamma$, CFG substantially improves sample quality, albeit with reduced diversity. Since its invention, CFG and its variants~\cite{hong2023improving, kynkaanniemi2024applying, karras2024guiding, sadatcads, sadat2024no, chung2024cfg++, ahn2025self} have become the backbone that powers the most advanced image generation systems~\cite{rombach2022high, saharia2022photorealistic, ramesh2022hierarchical}.

Despite practical success of CFG, its underlying mechanism remains largely unknown. As shown in~\cite{bradley2024classifier}, the CFG-perturbed reverse trajectory does not correspond to any known forward process, therefore, analyzing the effects of CFG requires case-by-case studies with explicit assumptions on the data distribution. For example, work~\cite{wutheoretical} proves that under an isotropic Gaussian mixture data assumption, CFG boosts classification accuracy at the cost of sample diversity. The work~\cite{chidambaramdoes} shows that under either 1-D mixtures of compactly supported distributions or 1-D isotropic Gaussian data assumptions, CFG guides the diffusion models towards sampling more heavily from the boundary of the support. Despite providing invaluable insights, these analyses rely on oversimplified assumptions that neglect critical aspects of real data, particularly the covariance structures of natural images. Consequently, it remains unclear how well these theoretical results generalize to diffusion models trained on complex image datasets.

In this work, we pursue a deeper understanding CFG's mechanism, focusing on two core questions: 
(i) \emph{What is the failure mode of naive conditional sampling, i.e., in what aspect is the generated images subpar compared to the training images?} and (ii) \emph{how does CFG mitigate this problem?}

To answer the first question, we show that the naive conditional suffers from a lack of class-specificity: images conditioned on different labels often share similar structures and lack distinct class features. We posit that this issue can be partially attributed to the covariance structures of different classes being insufficiently distinct. Recent studies~\cite{liunderstanding,wangunreasonable} observe that over a broad range of noise levels, diffusion models can be \textit{unreasonably} approximated by the optimal linear denoisers for the multivariate Gaussian distribution defined by the empirical mean and covariance of the training set. Consequently, the data covariance (and particularly its principal components, or PCs) heavily influences the generation. However, as we will demonstrate, different classes can share overly similar covariance structures, resulting in generated images that lack class-specific patterns.

Based on this intuition, we posit that CFG must identify the \emph{unique} features of the target class. To understand how this is achieved, we study the prototypical setting of the optimal \emph{linear} diffusion model, where we show that CFG guidance naturally decomposes into three components with distinct effects: 
(i) a \emph{mean-shift} term that approximately pushes the samples towards the direction of the class mean, (ii) a \emph{positive contrastive principal components (CPC)} term that enhances the target class’s unique features and (iii) a \emph{negative contrastive principal components (CPC)} term that suppresses the features prominent in the unconditional dataset. Despite the simplicity of the linear model, the linear CFG greatly improves the visual quality of generated samples in a way reminiscent of real-world, nonlinear deep diffusion models, implying that nonlinear CFG share a similar underlying working mechanism.
We then investigate how well the insights derived from the linear setting extend to actual diffusion models. We first show that at high to moderate noise levels, linear CFG yields highly similar effects as those of the nonlinear CFG. As noise decreases further and the diffusion model enters a highly nonlinear regime, the effects of linear CFG and actual nonlinear CFG begin to diverge.  Nevertheless, by interpreting denoising as weighted projection onto an adaptive basis, the insights from linear analysis can still shed light on the CFG's mechanism in the nonlinear regime.

\paragraph{Contributions.} Our main contributions are as follows:

\begin{itemize}[leftmargin=*]
    \item We identify the lack of class-specificity issue of naive conditional sampling, linking it to the non-distinctiveness of class covariances. Under a linear model assumption, we show CFG overcomes this issue by amplifying class-specific features, suppressing unconditional ones and shifting the samples in the direction of class mean.
    \item We validate these insights derived in the linear model on real diffusion models, demonstrating that:
    (i) at high to moderate noise levels, linear CFG closely matches the effects of nonlinear CFG, and
    (ii) at low noise levels, the insights from the linear analysis can still shed light on the mechanism of CFG in this nonlinear regime.
\end{itemize}
\vspace{-0.1in}

\section{Preliminaries}
\subsection{Optimal Linear Diffusion Model}
\label{sec: Gaussian inductive bias}
Suppose $p_\text{data}(\mb x)$ has mean $\mb \mu$ and covariance $\mb \Sigma$. Under the constraint that $\mathcal{D}(\mb x;\sigma (t))$ is a linear model (with a bias term), the optimal solution to~\eqref{denoising score matching} has the analytical form: 
\begin{align}
    \label{linear Gaussian diffusion model}
    \mathcal{D}_{\mathrm{L}}(\mb x;\sigma (t)) = \mb \mu+\mb U\Tilde{\mb \Lambda}_{\sigma(t)}\mb U^T (\mb x-\mb \mu),
\end{align}
where $\mb \Sigma=\mb U\mb\Lambda\mb U^T$ is the full SVD of the covariance matrix, $\mb \Lambda=\text{diag}(\lambda_1,\cdots,\lambda_d)$ is the singular values and $\Tilde{\mb \Lambda}_{\sigma(t)}=\diag \paren{\frac{\lambda_1}{\lambda_1+\sigma^2(t)}, \cdots,
     \frac{\lambda_d}{\lambda_d+\sigma^2(t)} }$. With this linear denoiser, the reverse diffusion ODE~\eqref{reverse ode} has the following closed-form expression (see~\cref{naive linear diffusion ode} for the proof):
 \begin{align}
    \label{Gaussian ODE}
     \mb x_t = \mb \mu + \sum_{i=1}^{d}\sqrt{\frac{\lambda_i+\sigma^2(t)}{\lambda_i+\sigma^2(T)}}\mb u_i^T(\mb x_T-\mb \mu)\mb u_i,
 \end{align}
where $T$ is the starting timestep and $\mb u_i$ is the $i^\text{th}$ singular vector of $\mb \Sigma$, which is also the $i^\text{th}$ principal component. Note that in this linear setting, the generated samples are largely determined by the data covariance. 

Recent studies~\cite{liunderstanding, wangunreasonable} show that for a wide range (high to moderate) of noise levels, deep network-based diffusion models can be well approximated by the linear model~\eqref{linear Gaussian diffusion model}, with $\mb \mu$ and $\mb \Sigma$ set to the empirical mean and covariance of the training data.
\begin{wrapfigure}[12]{r}{0.5\textwidth}
  \vspace{-0.5cm}
  \begin{center}
    \includegraphics[width=0.5\textwidth]{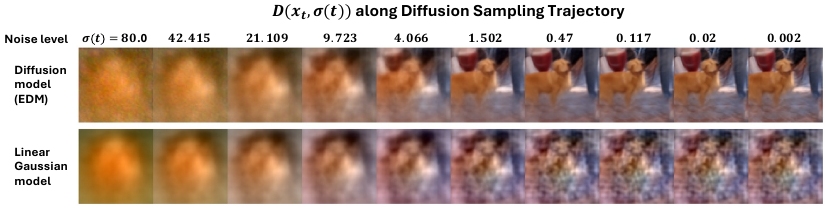}
  \end{center}
  \vspace{-0.3cm}
  \caption{\textbf{Comparison of Sampling Trajectories.} For high to moderate noise levels ($\sigma(t)\in(4,80]$), the linear denoisers well approximate the learned deep denoisers. Though the two models diverge in lower noise reigmes, their final samples still match in overall structure.}
  \label{fig:Sampling trajectory EDM v.s. Gaussian}
\end{wrapfigure}
As shown in~\Cref{fig:Sampling trajectory EDM v.s. Gaussian,fig:similarity between lienar and nonlinear models}, the sampling trajectories of the deep diffusion model (EDM) and the linear model share high similarity at high to moderate noise levels. Although the models begin to diverge at lower noise levels—where EDM exhibits strong nonlinearity and realistic image content begins to form—their final samples still share a similar overall structure. Moreover, as shown in~\cite{liunderstanding}, this similarity is particularly obvious when the deep network has limited capacity or the training is insufficient. Since $\mathcal{D}_\mathrm{L}(\mb x;\sigma(t))$ is the optimal denoiser for $p(\mb x;\sigma(t))$ induced by $p_\text{data}(\mb x)=\mathcal{N}(\mb \mu,\mb\Sigma)$, sampling with $\mathcal{D}_\mathrm{L}$ is equivalent to sampling from $\mathcal{N}(\mb\mu,\mb\Sigma)$. Hence, we refer to $\mathcal{D}_\mathrm{L}$ as the \emph{linear Gaussian model}.
\subsection{Contrastive Principal Component Analysis}
\label{contrastive principal component analysis}
Principal component analysis (PCA)~\cite{pearson1901liii, hotelling1933analysis} identifies directions that capture the most variances in a dataset. These principal components (PCs), which are equivalent to the singular vectors of the data covariance matrix, are widely used for data exploration and visualization. However, large variance alone does not guarantee that a PC captures the unique patterns tied to the dataset; it may instead reflect more general patterns such as foreground-background variations.

To discover low-dimensional structure that is unique to a dataset, the work~\cite{abid2018exploring} proposed the contrastive principal component analysis (CPCA), which utilizes a \emph{background} (or reference) dataset to highlight patterns unique to the \emph{target} dataset. Let $X$ and $Y$ be two datasets with covariance matrices $\mb \Sigma_X$ and $\mb \Sigma_Y$, respectively.
For a unit vector $\mb v\in \mathbb{S}^{d-1}$, its variances $\text{Var}_{X}(\mb v)$ and $\text{Var}_{Y}(\mb v)$ in the two datasets are:
\begin{align}
  \text{Var}_{X}(\mb v) &:= \mb v^T \mb\Sigma_{X} \mb v, 
  & \text{Var}_{Y}(\mb v) &:= \mb v^T \mb\Sigma_{Y} \mb v.
\end{align}
If $\mb v$ corresponds to a unique class-specific pattern of $X$, we expect $\text{Var}_{X}(\mb v)\gg\text{Var}_{Y}(\mb v)$,i.e., it explains significantly more variance in $X$ than in $Y$. Such directions, called the \emph{contrastive principal components (CPCs)}, can be found by maximizing:
\begin{align}
    \label{cpc objective}
    \argmax_{\mb v\in \mathbb{S}^{d-1}}\mb v^T(\mb \Sigma_{X}-\mb \Sigma_{Y})\mb v,
\end{align}
 which are essentially the top eigenvectors of $\mb \Sigma_{X}-\mb\Sigma_{Y}$. Geometrically, the first $k$ CPCs span the $k$-dimensional subspace that best fits the dataset $X$ while being as far as possible from $Y$ (see \cref{CPC appendix explaination} for details). Conversely, directions $\mb v$ for which $\text{Var}_{X}(\mb v)\approx\text{Var}_{Y}(\mb v)$ represent either universal structures shared by both $X$ and $Y$ or meaningless features lying in the null space of the data covariances—and are thus discarded as less interesting. Finally, a scalar factor can be introduced in~\eqref{cpc objective} to control the strength of the contrast.

\subsection{Posterior Data Covariance}
\label{posterior data covariance}
Consider $\mb x\sim p_\text{data}(\mb x)$ and $\mb x_t=\mb x+\sigma(t)\mb\epsilon$, where $\mb\epsilon\sim\mathcal{N}(\mb 0,\mb I)$. Then the posterior covariance of $p(\mb x|\mb x_t)$, denoted by $\text{Cov}[\mb x|\mb x_t]$, is proportional to the denoiser's Jacobian~\cite{manorposterior}:
\begin{align}
    \text{Cov}[\mb x|\mb x_t]=\sigma^2(t)\nabla\mathcal{D}(\mb x_t;\sigma(t)),
\end{align}
where $\nabla\mathcal{D}(\mb x_t;\sigma(t))=\frac{\partial\mathcal{D}(\mb x_t;\sigma(t))}{\partial\mb x_t}$ is the Jacobian of the optimal denoiser $\mathcal{D}(\mb x;\sigma(t))$ at input $\mb x_t$. Analogous to PCs, the singular vectors of $\text{Cov}[\mb x|\mb x_t]$ are the \emph{posterior PCs}, representing directions of maximal variances of all clean images that could have generated the noisy observation $\mb x_t$. In the case that $p_\text{data}=\mathcal{N}(\mb \mu,\mb\Sigma)$, we have $\text{Cov}[\mb x|\mb x_t]=\sigma^2(t)\mb U\Tilde{\mb\Lambda}_{\sigma(t)}\mb U^T$, matching $\nabla\mathcal{D}_{\mathrm{L}}(\mb x_t;\sigma(t))$, the Jacobians of the optimal linear denoiser~\eqref{linear Gaussian diffusion model} , and is independent of $\mb x_t$. In more general scenarios, one can approximate $\text{Cov}[\mb x|\mb x_t]$ by computing the network Jacobian at $\mb x_t$ via automatic differentiation.

\section{Analyzing CFG in Linear Model}
\label{Analyzing CFG with linear model theoretically}
\begin{figure*}[t]
  \centering
  \includegraphics[width=1\linewidth]{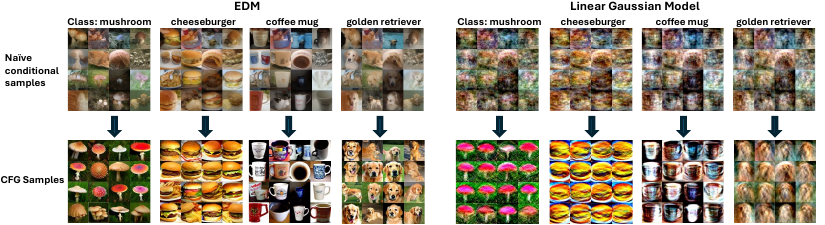}
  \caption{\textbf{Effects of CFG.} Left and right compare naive conditional sampling (top rows) versus CFG-guided sampling (bottom rows) for deep diffusion models (EDM) and linear Gaussian diffusion models, respectively. Each grid ceil corresponds to the same initial noise. While naive conditional samples lack class-specific clarity, CFG significantly improves both visual quality and distinctiveness. The conditional linear models are built with class-specific means and covariances. Please refer to~\cref{sec: more discussion on covariance structure} for more experiment results.}
  \label{fig:Gaussian inductive bias}
  \vspace{-0.2in}
\end{figure*}
In this section, we first show that naive conditional sampling often produces low-quality samples lacking clear class-specific features, which we attribute to the non-distinctiveness of class covariance matrices (\cref{naive conditional generation is sub-optimal}). We then theoretically analyze how CFG in the context of linear diffusion models alleviates this issue (\cref{how linear CFG leads to distinct samples}).
\subsection{Naive Conditional Generation Lacks Class-Specificity}
\label{naive conditional generation is sub-optimal}
\Cref{fig:Gaussian inductive bias}(left) (top row) shows the samples generated via naive conditional sampling \eqref{conditional sampling}. Qualitatively, these samples often exhibit poor image quality, with incoherent features that blend into the background and the class-specific image structures can be hard to discriminate. Moreover, even when conditioned on different class labels, images generated from the same initial noise share high structural similarity, suggesting that naive conditional sampling fails to capture discriminative, class-dependent patterns.

\begin{wrapfigure}[13]{r}{0.6\textwidth}
  \vspace{-0.5cm}
  \begin{center}
    \includegraphics[width=0.6\textwidth]{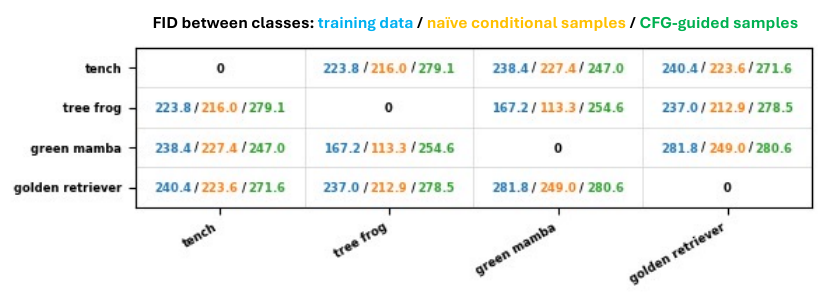}
  \end{center}
  \vspace{-0.4cm}
  \caption{\textbf{Class‑to‑Class Similarity.} Each cell reports the FID between datasets of two classes, built with (i) training data (ii) data generated by naive conditional sampling and (iii) data generated by CFG sampling (refer to~\cref{subsec: Quantitative Results} for experiment details and more results.)}
  \label{fig:FID for training data naive conditional generated data and cfg data}
\end{wrapfigure}

To quantify this loss of class-specificity, we compute the pairwise inter-class similarity with the FID metric~\cite{heusel2017gans}. For each pair of classes, we construct two datasets $X$ and $Y$ and evaluate the FID between them. As shown in~\Cref{fig:FID for training data naive conditional generated data and cfg data}, when $X$ and $Y$ are built with images generated with naive conditional sampling, the FID (colored in orange) is consistently lower than when they are built with the training data (colored in blue). Since lower FID indicates higher similarity, this result confirms that compared with the training images, which represent the ground truth data distribution, images generated by naive conditional sampling are less distinguishable across classes.

This issue is especially pronounced in linear diffusion models. As shown in \Cref{fig:Gaussian inductive bias}(right, top row), samples generated with linear diffusion models built with class-specific means and covariances appear highly similar. From \eqref{Gaussian ODE}, we see that the linear sampling trajectory is governed by the data covariance: $\mb x_t$ is a linear combination of PCs, weighted by (i) the correlation $\mb u_i^T(\mb x_T-\mb\mu)$ between the mean-subtracted initial noise $\mb x_T$ and the $i$-th PC, and (ii) scaling factors $\sqrt{\frac{\lambda_i+\sigma^2(t)}{\lambda_i+\sigma^2(T)}}$ that emphasize leading PCs. Consequently, if class-conditional covariances lack sufficiently discriminative structures (which is indeed true as shown in~\cref{subsec: Covariances matrices are not distinct enough}), generated samples will appear similar regardless of class label. This lack of class-specificity aligns with prior findings~\cite{abid2018exploring}, which shows that PCs often capture generic image variations (e.g., foreground-background), rather than class-specific patterns.

The existence of the class-specificity gap implies these models fail to fully capture the higher-order statistics of the training data: if they did, naive conditional sampling, which by 
construction samples from the target conditional distribution, would already produce high quality samples, and CFG would only distort the target distribution. Linear diffusion models represent an extreme case: due to the linear constraint, they can only learn the first and second-order moments (mean and covariance) of the training data, which despite being fundamental data statistics, cannot capture the rich, nonlinear dependencies necessary for realistic generation. In particular, when covariances across classes share high similarity, samples initialized from the same noise become visually alike regardless of label. 

We hypothesize that real-world diffusion models inherit similar limitations. Although nonlinear diffusion models surely learn beyond second-order statistics, as discussed in~\cref{sec: Gaussian inductive bias}, for high to moderate noise levels, they can be well approximated by linear models, especially under limited model capacity or insufficient training. Indeed,~\Cref{fig:Gaussian inductive bias} (top row) and \Cref{fig:various_steps_samples_visualization}, \ref{fig:similarity between lienar and nonlinear models} demonstrate that linear models reproduce the coarse-grained structures of nonlinear diffusion samples, implying that the covariance structure plays a significant role in shaping the high-level features of the generated samples. These observations reflect a well-known \textit{simplicity bias}, where deep networks favor learning low-order, linearly structured representations over complex, higher-order dependencies~\cite{kalimeris2019sgd}. Hence, if the covariances are indistinct across classes, sample quality can be limited even in nonlinear models (see~\cref{subsec: Covariances matrices are not distinct enough} for more discussion).

As quantitatively shown in~\Cref{fig:FID for training data naive conditional generated data and cfg data}, CFG significantly increases the inter‑class separation: FID (colored in green) between different generated classes rises. Qualitatively,~\Cref{fig:Gaussian inductive bias} (bottom row) shows that CFG substantially improves both linear and nonlinear models, producing visibly better samples with enhanced class-specific structures. Similar effects of CFG across both linear and nonlinear models motivate us to use a linear model as a simplified prototype to analyze how CFG reshapes the generation process and why it is effective.

\subsection{How Linear CFG Leads to Distinct Generations}
\label{how linear CFG leads to distinct samples}
We now dissect how CFG, in the linear diffusion models, produces samples with distinct class-specific features. Consider two independent optimal linear denoisers, $\mathcal{D}_{\mathrm{L}}(\mb x_t;\sigma(t),\mb c)$ for conditional data and $\mathcal{D_\mathrm{L}}(\mb x_t;\sigma(t))$ for unconditional data, with means $\mb\mu_c$, $\mb\mu_{uc}$ and covariances $\mb\Sigma_c=\mb U_c\mb\Lambda_c\mb U_c^T$ and $\mb\Sigma_{uc}=\mb U_{uc}\mb\Lambda_{uc}\mb U_{uc}^T$ , respectively. Substituting the optimal linear denoiser~\eqref{linear Gaussian diffusion model} into \eqref{cfg sampling}, the CFG-guided sampling process can be decomposed into three terms: 
\begin{align}
    d\mb x_t &= -\sigma(t)\bigl(f_{c}(\mb x_t,t) + g_{cpc}(\mb x_t,t) + g_{mean}(t)\bigr)dt\label{all},
\end{align}
where by letting $\Tilde{\mb \Sigma}_{c,t} = \mb U_c\Tilde{\mb\Lambda}_{\sigma(t),c}\mb U_c^T$ and $\Tilde{\mb \Sigma}_{uc,t} = \mb U_{uc}\Tilde{\mb\Lambda}_{\sigma(t),uc}\mb U_{uc}^T$, each term takes the following form: $(i)~f_{c}(\mb x_t,t) = \frac{1}{\sigma^2(t)}( \tilde{\mb\Sigma}_{c,t}-\mb I)(\mb x_t-\mb\mu_c)$, $(ii)~g_{cpc}(\mb x_t,t) = \frac{\gamma}{\sigma^2(t)}(\Tilde{\mb \Sigma}_{c,t} - \Tilde{\mb \Sigma}_{uc,t})(\mb x_t-\mb\mu_c)$, and $(iii)~g_{mean}(t) = \frac{\gamma}{\sigma^2(t)}(\mb I-\Tilde{\mb \Sigma}_{uc,t} )(\mb \mu_c-\mb \mu_{uc})$.

Here, $f_c(\mb x_t, t)$ is the standard conditional score, and $g_{cpc}(\mb x_t,t)$ plus $g_{mean}(t)$ form the CFG guidance (derivation for the decomposition is provided in~\cref{CFG guidaed ode proof}). Let $\mb V_{\sigma(t)}\hat{\mb\Lambda}_{\sigma(t)}\mb V_{\sigma(t)}^T$ be the eigen decomposition of 
$\Tilde{\mb \Sigma}_{c,t} - \Tilde{\mb \Sigma}_{uc,t}$, whose spectrum contains both positive and negative eigenvalues (see~\Cref{fig:eigen spectrum}), $g_{cpc}(\mb x_t, t)$  can be split accordingly into \emph{positive} and \emph{negative} CPC components:
\begin{align}
    \underbrace{\frac{\gamma}{\sigma^2(t)}(\mb V_{\sigma(t),+}\hat{\mb\Lambda}_{\sigma(t),+}\mb V^T_{\sigma(t),+})(\mb x_t-\mb\mu_c)}_\text{positive CPC guidance}=\frac{\gamma}{\sigma^2(t)}\sum_i \hat{\lambda}_{+,i}\mb v_{+,i} \bigl(\mb v_{+,i}^T(\mb x_t-\mb\mu_c)\bigl),    \label{scaled positive cpc subspace projection}\\ 
     \underbrace{\frac{\gamma}{\sigma^2(t)}(\mb V_{\sigma(t),-}\hat{\mb\Lambda}_{\sigma(t),-}\mb V^T_{\sigma(t),-})(\mb x_t-\mb\mu_c)}_\text{negative CPC guidance}=\frac{\gamma}{\sigma^2(t)}\sum_i \hat{\lambda}_{-,i}\mb v_{-,i} \bigl(\mb v_{-,i}^T(\mb x_t-\mb\mu_c)\bigl),    \label{scaled negative cpc subspace projection}
\end{align}
where $\mb V_{\sigma(t),+}$ and $\mb V_{\sigma(t),-}$ contain eigenvectors $\mb v_{+,i}$ and $\mb v_{-.i}$ corresponding to positive and negative eigenvalues $\hat{\mb\Lambda}_{\sigma(t),+}$ and $\hat{\mb\Lambda}_{\sigma(t),-}$ respectively. As discussed in \cref{posterior data covariance}, $\Tilde{\mb \Sigma}_{c,t}$ and $\Tilde{\mb \Sigma}_{uc,t}$ are up to a scaling factor $\sigma^2(t)$ equivalent to the conditional and unconditional posterior covariances of $p_\text{data}(\mb x|\mb c)=\mathcal{N}(\mb \mu_c,\mb\Sigma_c)$ and $p_\text{data}(\mb x)=\mathcal{N}(\mb \mu_{uc}, \mb \Sigma_{uc})$. Hence, $\mb V_{\sigma(t)}$ are the CPCs which contrast between $X\sim p_{data}(\mb x|\mb x_t,\mb c)$ and $Y\sim p_\text{data}(\mb x|\mb x_t)$. Specifically, $\mb{V}_{\sigma(t),+}$ captures directions of higher conditional variance (class-specific features), while $\mb{V}_{\sigma(t),-}$ captures directions of higher unconditional variance (features more prevalent in the unconditional data). 

\begin{figure}[t!]
    \centering
    \includegraphics[width=1\linewidth]{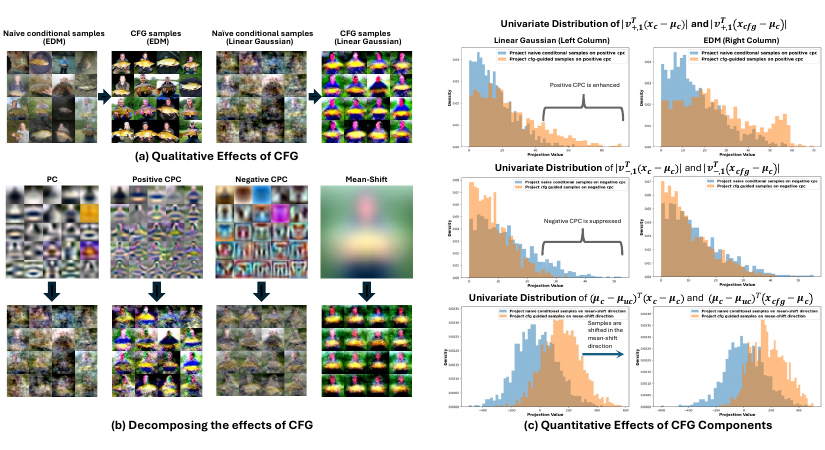}
      \vspace{-0.2in}
    \caption{\textbf{Distinct effects of different CFG components.}
    (a) CFG substantially enhances class-specific features (in both EDM and linear diffusion). 
    (b) Top row: PCs, positive/negative CPCs, and \(\boldsymbol\mu_c - \boldsymbol\mu_{uc}\). Bottom row: generated samples when each component is applied in isolation. (c) One‑dimensional densities of generated samples after projection onto key directions. The left column corresponds to the linear diffusion model, whereas the right column corresponds to the EDM model. Top row: project onto leading positive CPC. Middle row: project onto negative CPC. Third row: project onto the mean-shift direction. Here we only plot the resulting histograms for the first positive and negative CPCs but the same patterns hold for subsequent CPCs. For experimental details and more results, please refer to~\cref{Distinct Effects of the CFG Components appendix}.}
    \label{distinct effects of CFG on linear models}
    \vspace{-0.1in}
\end{figure}

\textbf{Distinctive Effects of the CFG Components.}
\Cref{distinct effects of CFG on linear models}(a) shows that for both nonlinear (EDM) and linear models, CFG significantly enhances the characteristic pattern—a person holding a fish—of the "tench" class from ImageNet~\cite{deng2009imagenet}. Next, we isolate the roles of each CFG term by selectively enabling only one at a time within the linear model. In the following discussion, we omit the negative sign in~\eqref{all} since the ODE runs backward in time:
\begin{itemize}[leftmargin=*]
\item The \textbf{positive CPC term}~\eqref{scaled positive cpc subspace projection} projects $\mb x_t - \mb\mu_c$ onto the subspace spanned by the positive CPCs, i.e., the eigenvectors $\mb{v}_{+,i}$ associated with positive eigenvalues, with each component scaled by its eigenvalue $\hat{\lambda}_{+,i}$ and the guidance strength $\gamma$. Since $\hat{\lambda}_{+,i}\geq 0$,~\eqref{scaled positive cpc subspace projection} is added to $\mb x_t$, i.e., the components of $\mb x_t-\mb\mu_c$ that align with the positive CPCs, which represent the class-specific features, are amplified. 
~\Cref{distinct effects of CFG on linear models} (b) (second column) show the first 25 positive CPCs of $\mb \Sigma_c-\mb \Sigma_{uc}$ \footnote{
Although $\mb V_{\sigma(t)}$ depends on $\sigma(t)$, over a wide range of noise levels (especially high ones), it remains close to the eigenvectors of $\mb\Sigma_{c}-\mb\Sigma_{uc}$. We provide its full evolution across time in~\Cref{fig: CPC_vs_PC_evolution}.} and the resulting samples. Compared to the conditional PCs of the dataset, the positive CPCs better capture the unique patterns of the class, which emerge visibly in the generated images.
\item Similarly, the \textbf{negative CPC term}~\eqref{scaled negative cpc subspace projection} projects $\mb x_t - \mb{\mu}_c$ onto the negative CPC directions $\mb{v}_{-,i}$.
Since $\hat{\lambda}_{-,i}<0$, these components are subtracted from $\mb x_t$, suppressing features associated more strongly with the unconditional data.~\Cref{distinct effects of CFG on linear models} (b) (third column) shows the first 25 negative CPCs and the resulting generations. Although visually less interpretable, these directions represent common but target-class-irrelevant features in the unconditional data. Suppressing them reduces background clutter and irrelevant content, making class-relevant structures more salient.

\item In the context of linear diffusion model, it can be shown that (see proof in~\cref{CFG guidaed ode proof}):
\begin{align}
    g_{mean}(t)=\gamma\mathbb{E}_{\mb x\sim p(\mb x|\mb c;\sigma(t))}[\nabla_{\mb x}\log p(\mb x|\mb c;\sigma(t))-\nabla_{\mb x}\log p(\mb x;\sigma(t))].
\end{align}
Thus, the \textbf{Mean-shift term $g_{mean}(t)$} can be interpreted as the probability-weighted average of the steepest ascent direction that maximizes the difference (log-likelihood ratio) of the noise-mollified conditional and unconditional distributions. Note that when $\sigma(t)$ is large, $g_{mean}(t)$ approximately shifts $\mb x_t$ in the direction of $\mb\mu_c-\mb\mu_{uc}$, i.e., the difference between conditional and unconditional mean, since $\mb I-\tilde{\mb\Sigma}_{uc,t}\approx\mb I$. 
As \(\sigma(t)\) decreases,  $\mb I-\tilde{\mb\Sigma}_{uc,t}$ progressively shrink the components $\mb\mu_c-\mb\mu_{uc}$ lying in the column space of $\mb U_{uc}$ (the covariances of image datasets are typically low-rank), while preserving its energy in the null space.

\Cref{distinct effects of CFG on linear models}(b), fourth column, shows that the mean-shift term enhances the structure of class mean in the generated samples. However, unlike the positive CPC term, $g_{mean}(t)$ is independent of $\mb x_t$, thus producing more homogeneous samples with reduced diversity.
\end{itemize}

\textbf{Analytical Solution to the CFG Trajectory.}
To better understand the distinct effects of the CFG components, we aim to examine the global solution to the linear ODE system~\eqref{all}. However, the variables in the general solution of~\eqref{all} are coupled and difficult to interpret. To obtain a more tractable expression, we follow~\cite{flury1984common,10.5555/74760} to make the following assumption:

\begin{assum}\label{assum}
    \label{main text Common Principal Components Assumption}
    The covariance matrices $\mb\Sigma_c$ and $\mb\Sigma_{uc}$ are simultaneously diagonalizable, i.e., $\mb\Sigma_{uc}=\mb U_c\mb \Lambda_{uc}\mb U_c^T$, where $\mb U_c\in \mathbb{R}^d$ are the singular vectors (principal components) of the conditional covariance. Here $\mb\Lambda_{uc}$ is not necessarily ordered by the magnitude of the singular values.  
\end{assum}
Assumption \ref{assum}, known as the \textit{Common Principal Components Assumption} is widely applied to analyze structural relationships across data groups. Under this assumption, the relative importance of the $i$-th principal component $\mb u_{c,i}$ in the conditional and unconditional datasets is determined by the relative magnitudes of its associated singular values.  If $\lambda_{c,i} > \lambda_{uc,i}$, then $\mb{u}_{c,i}$ is the positive CPC as it captures more variance in the conditional distribution, while $\lambda_{c,i} < \lambda_{uc,i}$ implies that $\mb{u}_{c,i}$ is more relevant to the unconditional distribution; therefore, it is a negative CPC. 
\begin{theorem}
\label{theorem}
Under Assumption \ref{assum}, the solution to the linear CFG process~\eqref{all} is:
\begin{small}
\begin{align}
    \mb x_t&=  \mb\mu_c + \sum_{i=1}^d\textcolor{blue}{h(\lambda_{c,i},\lambda_{uc,i})^{\frac{\gamma}{2}}} \sqrt{\frac{\lambda_{c,i}+\sigma^2(t)}{\lambda_{c,i}+\sigma^2(T)}}\mb u_{c,i}^T(\mb x_T-\mb \mu_c) \mb u_{c,i}\nonumber + \textcolor{red}{\gamma\mb U_c\mb B_{\sigma{(t)}}\mb U_c^T(\mb\mu_c-\mb\mu_{uc})}\label{constant perturbation}, 
\end{align}
\end{small}
where $h(\lambda_{c,i},\lambda_{uc,i})=\frac{\lambda_{c,i}+\sigma^2(t)}{\lambda_{c,i}+\sigma^2(T)}\cdot\frac{\lambda_{uc,i}+\sigma^2(T)}{\lambda_{uc,i}+\sigma^2(t)}$ and $\mb B_{\sigma_t}=\text{diag}(b_{\sigma(t),1},...,b_{\sigma(t),d})$ has diagonal entries $b_{\sigma(t),i}$ depending only on $\lambda_{uc,i},\lambda_{c,i}$ and $\sigma(t)$. 
\end{theorem}

The proof is postponed to~\cref{sec:proof}. Compared to the solution of the standard conditional sampling \eqref{Gaussian ODE}, the CFG guidance introduces the following two effects:
\begin{itemize}[leftmargin=*]
\item \textbf{CPC guidance $g_{cpc}(\mb x_t,t)$} introduces an additional scaling factor $h(\lambda_{c,i},\lambda_{uc,i})^\frac{\gamma}{2}$ for each component $\mb u_{c,i}$ of $\mb x_t$. Since $h(\lambda_{c,i},\lambda_{uc,i})\geq 1$ only if $\lambda_{c,i}\geq \lambda_{uc,i}$, the positive CPCs are enhanced. Conversely, the negative CPCs are suppressed. The guidance strength $\gamma$ serves as an additional control over the degree of enhancement or suppression. 

\item \textbf{Mean-shift guidance term $g_{mean}(t)$} shifts $\mb x_t$ by $\gamma\mb U_c\mb B_{\sigma{(t)}}\mb U_c^T(\mb\mu_c-\mb\mu_{uc})$, a direction determined by the class-conditional mean offset $\mb\mu_c-\mb\mu_{uc}$. Crucially, this shift is independent of the initial noise $\mb{x}_T$ (and intermediate state $\mb x_t$) and is thus applied consistently to all samples, promoting canonical class features but reducing diversity.
\end{itemize}
\paragraph{Empirical Verification.}
In~\Cref{subsec: empirical verification}, we provide an empirical validation of~\Cref{theorem} using a 2D synthetic dataset that satisfies Assumption \ref{assum}.
Here, we further verify the CFG's effects of enhancing (suppressing) CPC components and shifting samples towards the mean-shift direction in natural image dataset through the following experiment:
\begin{itemize}[leftmargin=*]
    \item For a chosen class, generate 1,000 samples using naive conditional sampling (denoted by $\mb x_c$) and 1,000 samples using CFG (denoted by $\mb x_\text{cfg}$), and center both sets by subtracting the class mean $\mb \mu_c$.
    \item Project each sample onto the positive CPCs (denoted as $\mb v_{+}$), the negative CPCs (denoted by $\mb v_{-}$), and the mean-shift vector (denoted by $\mb\mu_c-\mb\mu_{uc}$) to obtain a series of univariate distributions.
\end{itemize}
The above experiments are conducted on both linear and nonlinear (EDM) diffusion models. The resulting univariate distributions are shown in~\Cref{distinct effects of CFG on linear models}(c). Compared with naive conditional sampling, CFG shifts probability mass toward higher projection values along the positive-CPC and mean-shift directions, and toward lower values along the negative-CPC direction, indicating that the first two are amplified whereas the third is suppressed.

\section{Investigating CFG in Nonlinear Models}
\label{investigate CFG in nonlinear models}
\begin{figure*}[t!]
    \centering
    \includegraphics[width=0.9\linewidth]{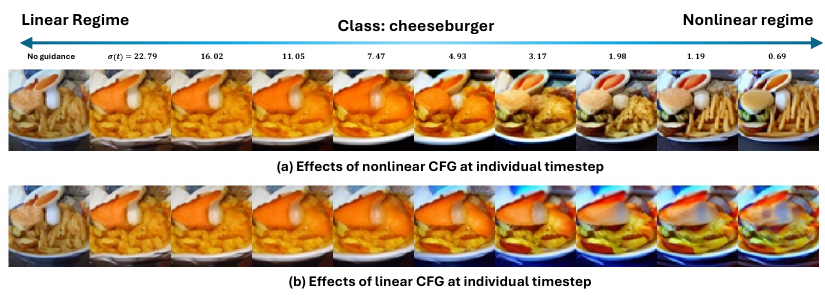}
    \caption{\textbf{Linear-to-nonlinear transition in diffusion models.} (a) and (b) compare nonlinear CFG and linear CFG applied to a deep diffusion model (EDM). The leftmost column shows unguided samples; subsequent columns show final samples when guidance is applied only at a specific noise level, with $\gamma=15$ (See~\Cref{fig:linear_nonlinear_transition_extra} for more examples).}
   \label{fig:linear_nonlinear_transition}
    \vspace{-0.2in}
\end{figure*}

We now explore how the findings from the linear analysis extend to real-world diffusion models. Recent studies~\cite{raya2024spontaneous,liunderstanding,wangunreasonable} show that diffusion models transition from a linear regime to a nonlinear regime as the noise level decreases. In the linear regime, where $\sigma(t)$ is large, the learned diffusion denoisers $\mathcal{D}_{\mb\theta}$ can be well approximated by the optimal linear denoiser $\mathcal{D}_\mathrm{L}$~\eqref{linear Gaussian diffusion model} (see both qualitative and quantitative verification in~\cref{appendix linear to nonlinear transition}). As $\sigma(t)$ decreases, the diffusion model enters the nonlinear regime where $\mathcal{D}_{\mb\theta}$ diverges from $\mathcal{D}_\mathrm{L}$. Interestingly, this linear-to-nonlinear transition correlates with the coarse-to-fine effects of CFG. As shown in~\Cref{fig:linear_nonlinear_transition,fig:linear_nonlinear_transition_extra}, in the linear regime, linear and nonlinear CFG produce similar effects, substantially reshaping the global structure of the samples. In contrast, in the nonlinear regime, nonlinear CFG primarily refines local details while preserving the overall structure, leading to different effects as those of linear CFG. This linear-to-nonlinear, coarse-to-fine transition motivates our separate analyses of CFG behavior in each regime.

\subsection{CFG in the Linear Regime}
\label{linear regime}
\begin{wrapfigure}[15]{r}{0.6\textwidth}
  \begin{center}
    \includegraphics[width=0.6\textwidth]{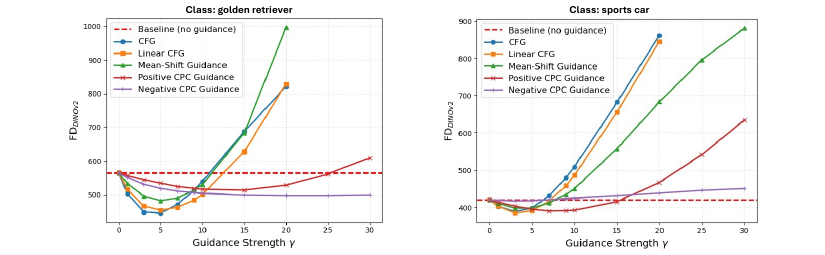}
  \end{center}
  \caption{\textbf{FD\textsubscript{DINOv2} Scores.} The scores are computed over 50,000 samples. The reported values are relatively high because the scores are computed separately per class, which often has limited number of training images. It is well known that FD\textsubscript{DINOv2} scores can appear inflated when the reference dataset size is small.}
   \label{fig:FD_score_main_text}
\end{wrapfigure}

\begin{figure*}[t!]
    \centering
    \includegraphics[width=01\linewidth]{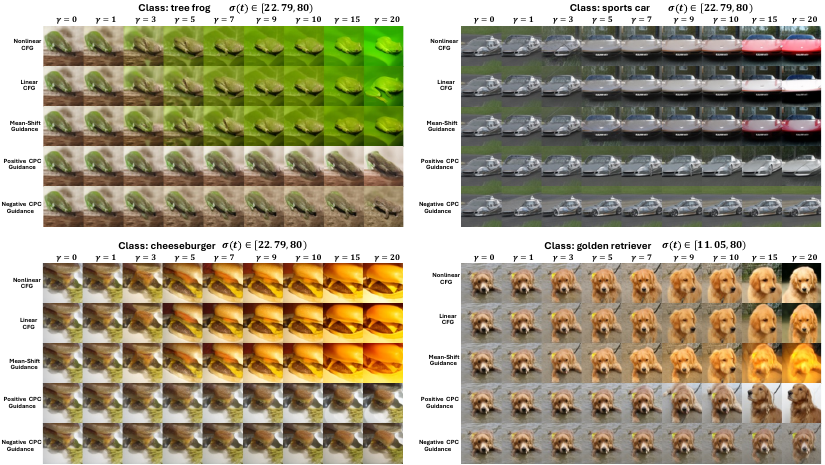}
    \caption{\textbf{Effects of CFG in Linear Regime.} Each row demonstrates the impact of different guidance types applied to EDM within the linear regime, with varying guidance strength $\gamma$. The guidance is applied only within intervals specified in the subtitles, where the model exhibits linear behavior. For additional experimental results, please refer to \cref{CFG in Nonlinear deep diffusion model appendix}.}
    \label{fig:CFG_linear_regime}
    \vspace{-0.2in}
\end{figure*}

\Cref{fig:CFG_linear_regime,fig:FD_score_main_text} illustrates the effects of separately applying (i) nonlinear CFG, (ii) linear CFG, (iii) mean-shift guidance, (iv) positive CPC guidance, and (v) negative CPC guidance within the linear regime of EDM over a broad range of $\gamma$. As expected, linear CFG\footnote{Note that the “linear CFG” here differs from the “linear CFG” in \cref{Analyzing CFG with linear model theoretically}, where both the naive conditional score and the cfg guidance are linear. In contrast, the linear guidance in this section, along with its components, is applied to a real-world \emph{deep} diffusion model.} produces results that closely match those of nonlinear CFG, both significantly altering the overall structures of unguided samples. Notably, decomposing linear CFG provides further insights:

\textbf{Mean-shift guidance dominates CFG in the linear regime.} As shown in \Cref{fig:CFG_linear_regime}, qualitatively, mean-shift guidance alone replicates the effects of both linear and nonlinear CFG. Consistent with this observation, FD\textsubscript{DINOv2} scores confirm that the mean-shift term is the main contributor to CFG’s overall behavior. Because mean-shift term is independent of the sampling trajectory, it reduces sample diversity. As shown in~\Cref{fig:FD_score_main_text}, mean-shift guidance improves generation quality only within a limited range of $\gamma$, after which further increases in $\gamma$ degrade FD\textsubscript{DINOv2} scores, reflecting a loss of diversity.

Moreover, the observation that the sample-independent mean-shift guidance alone leads to improved FD\textsubscript{DINOv2} score implies that simply initializing the sampling process from a mean-shifted Gaussian, $\mb x_T\sim \mathcal{N}(\gamma(\mb\mu_c-\mb\mu_{uc}),\sigma^2(T)\mb I)$, with no additional guidance applied, can improve the generation quality, which is verified in~\cref{Mean-shifted noise initialization trick}. Practically, this initialization trick avoids the per-sample, per-timestep network inference required by CFG, hence could be beneficial in applications where inference speed is important. Theoretically, the observation that the mean-shifted initialization yields better sample quality compared to the commonly used zero-mean initialization suggests the existence of class-specific clusters in the intermediate noisy distribution $p(\mb x;\sigma(t))$. Understanding how these clusters are formed and what additional structures they possess is an interesting future direction.

\textbf{CPC guidance also improves generation quality.} Although overshadowed by the mean-shift term, applying CPC guidance independently offers notable benefits as well. Qualitatively, positive and negative CPC terms preserve the global structure of unguided samples while refining existing features, remaining effective over a broader range of $\gamma$. Moreover, CPC guidance sometimes mitigate the artifacts introduced by the mean-shift term, such as color oversaturation in the golden retriever example at $\gamma=20$. Lastly, we note that the effects of CPC guidance can vary by class. As shown in~\Cref{fig:FD_score_main_text}, negative CPC term improves FD\textsubscript{DINOv2} scores for "golden retriever" but has minimal effect on "sports car". These findings are verified on 10 classes, with additional results presented in~\cref{CFG linear reimge discussion appendix}, \Cref{fig:linear_regime_qualitative_extra,fig:linear_regime_qualitative_extra_2,fig:linear_regime_quantatitive_extra,fig:FD_score}.
\subsection{CFG in the Nonlinear Regime}
\label{sec: nonlienar regime investigation}
In the nonlinear regime where $\sigma(t)$ is small, as shown in~\Cref{fig:linear_nonlinear_transition}(b), the effects of linear CFG diverge from those of the actual nonlinear CFG. By Tweedie's formula, the CFG guidance~\eqref{cfg guidance term} can be expressed as $g(\mb x,t)=\frac{\mathcal{D}(\mb x;\sigma(t),\mb c)-\mathcal{D}(\mb x;\sigma(t)}{\sigma^2(t)}$, where $\mathcal{D}(\mb x;\sigma(t),\mb c)$ and $\mathcal{D}(\mb x;\sigma(t))$ denote the optimal conditional and unconditional denoisers minimizing ~\eqref{denoising score matching}. Unlike in the linear setting, these denoisers do not admit closed-form expressions in the nonlinear regime, making analytical study difficult. Nevertheless, when denoisers are parameterized by deep networks with no additive 'bias' parameters, their input-output mappings are locally piecewise linear~\cite{kadkhodaiegeneralization,mohan2020robust}, satisfying:
\begin{equation}
\label{eq:jacobian-piecewise-linear}
\mathcal{D}\bigl(\mb x;\sigma(t),\mb c\bigr)
  = \nabla_{\mb x}\mathcal{D}\bigl(\mb x;\sigma(t),\mb c\bigr)\,\mb x,
\qquad
\mathcal{D}\bigl(\mb x;\sigma(t)\bigr)
  = \nabla_{\mb x}\mathcal{D}\bigl(\mb x;\sigma(t)\bigr)\,\mb x,
\end{equation}
 where $\nabla_{\mb x}\mathcal{D}(\mb x;\sigma(t),\mb c)$ and $\nabla_{\mb x}\mathcal{D}(\mb x;\sigma(t))$ are the local Jacobians of the denoisers. In this case, CFG guidance becomes \( \frac{(\nabla_{\mb{x}} \mathcal{D}(\mb{x}; \sigma(t), \mb{c}) - \nabla_{\mb{x}} \mathcal{D}(\mb{x}; \sigma(t))) \mb{x}}{\sigma^2(t)} \) 
, which shares a similar form as $g_{cpc}(\mb x, t)$ defined under the linear setting in~\eqref{all} since $\Tilde{\mb \Sigma}_{c,t} - \Tilde{\mb \Sigma}_{uc,t}=\nabla_{\mb x}\mathcal{D}_{\mathrm{L}}(\mb x;\sigma(t),\mb c)-\nabla_{\mb x}\mathcal{D}_{\mathrm{L}}(\mb x;\sigma(t))$, where $\mathcal{D}_{\mathrm{L}}$ is the optimal linear denoiser. Thus, the guidance can again be decomposed into positive and negative CPC components, enhancing the former and suppressing the latter. The key distinction from the linear setting is that here, the CPCs are adaptive to $\mb x$.

The bias-free denoisers belong to the broader class of pseudo-linear denoisers~\cite{milanfar2024denoising,romano2017little}, which admit the form $\mathcal{D}(\mb x;\sigma(t))=\mb W(\mb x;\sigma(t))\mb x$, where $\mb W(\mb x;\sigma(t))$ is symmetric and input-dependent. Importantly, it is shown in~\cite{milanfar2024denoising} that if the origin is a stationary point of the log-density, i.e., $\nabla_{\mb x}\log p(\mb x;\sigma(t))|_{\mb x=\mb 0}=\mb 0$, then the optimal denoiser must possess such a piecewise linear structure. Even if the diffusion models are not bias-free and the locally linear property does not hold exactly,~\eqref{eq:jacobian-piecewise-linear} still serves as a reasonable proxy. As discussed in~\cref{posterior data covariance}, the Jacobian $\nabla_{\mb x}\mathcal{D}(\mb x;\sigma(t))$ is proportional to the posterior covariance. Its leading singular vectors capture the dominant structures shared by all plausible clean images corresponding to the noisy input $\mb x$, while directions associated with near-zero singular values span a null space irrelevant to the image content. Hence,~\eqref{eq:jacobian-piecewise-linear} performs a weighted projection onto the subspace encoding the most informative image structures—effectively functioning as a valid denoising operator.
\begin{figure*}[t!]
    \centering
    \includegraphics[width=1\linewidth]{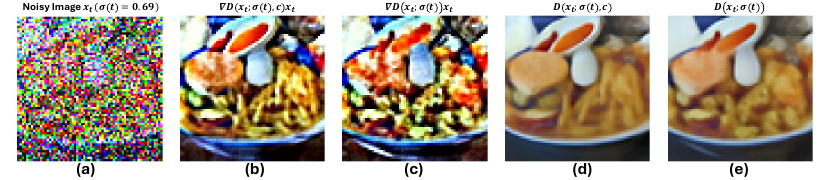}
\caption{\textbf{Denoising Results.} (a) Noisy input image. (b)–(e) show the denoised outputs generated with (i) conditional Jacobian, (ii) unconditional Jacobian, (iii) actual conditional denoiser, and (iv) actual unconditional denoiser, respectively.}
   \label{fig:Jacobian denoiser}
    \vspace{-0.2in}
\end{figure*}
Indeed, as shown in~\Cref{fig:Jacobian denoiser}(b)--(c), both conditional and unconditional Jacobians effectively denoise the input, although their outputs appear brighter and sharper than those from the actual denoisers in (d)–(e). Comparing ~\Cref{fig:Jacobian denoiser}(b) and (c), we find that the conditional and unconditional Jacobians yield denoised outputs with similar global structure, which implies both capture the generic structure of the current sample. However, the conditional Jacobian additionally preserves finer, class-specific details. A similar pattern holds for the actual denoisers shown in \Cref{fig:Jacobian denoiser}(d) and (e). 

\begin{figure*}[h]
    \centering
    \includegraphics[width=1\linewidth]{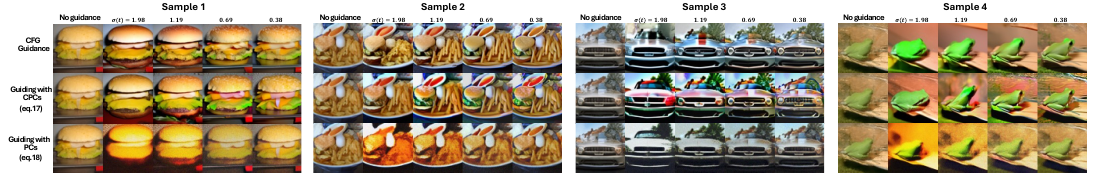}
    \caption{\textbf{Effects of CFG in the Nonlinear Regime.} Different guidance methods, each with a fixed strength of $\gamma=15$, are applied at individual timesteps in the nonlinear regime. Each image shows the final output when guidance is applied solely at the timestep indicated at the top. Note that \eqref{scaled positive cpc subspace projection Jacobian} matches the effects of CFG by enhancing finer image details, whereas \eqref{scaled positive PCA projection Jacobian} does not improve generation quality. For additional experimental results, please refer to \Cref{fig:construct_nonlinear_CFG_extra}.} 
    \label{fig:guiding effects in nonlinear regime}
    \vspace{-0.1in}
\end{figure*}

For guidance purposes, our goal is to selectively enhance these fine, class-specific details that the conditional denoiser captures but the unconditional one does not. Achieving this requires identifying directions that encode class-dependent information from those represent generic structures. Empirically, as shown in~\Cref{fig:guiding effects in nonlinear regime}, the following guidance, inspired by the positive CPC guidance~\eqref{scaled positive cpc subspace projection}, can lead to similar effects as the actual nonlinear CFG, sharpening image details:
\begin{align}
    \label{scaled positive cpc subspace projection Jacobian}
    \frac{\gamma}{\sigma^2(t)}\sum_i \hat{\lambda}_{+,i}\,\mb v_{+,i}\,\bigl(\mb v_{+,i}^T \mathcal{D_{\mb\theta}}(\mb x_t;\sigma(t),\mb c)\bigl),
\end{align}
where $\hat{\lambda}_{+,i}$ and $\mb v_{+,i}$ denote the positive eigenvalues and eigenvectors of $\nabla\mathcal{D}_{\mb\theta}(\mb x_t;\sigma(t),\mb c) -\nabla\mathcal{D}_{\mb\theta}(\mb x_t;\sigma(t))$.
Unlike~\eqref{scaled positive cpc subspace projection}, this guidance applies projection to the denoiser's output rather than the noisy input $\mb x_t$, which we find leads to better qualitative results. For comparison, we also test the following non-selective guidance which enhances all the conditional posterior PCs:
\begin{align}
    \label{scaled positive PCA projection Jacobian}
    \frac{\gamma}{\sigma^2(t)}\sum_i \lambda_{c,i}\,\mb u_{c,i}\,\bigl(\mb u_{c,i}^T \mathcal{D}_{\mb\theta}(\mb x_t;\sigma(t),\mb c)\bigl),
\end{align}
where $\lambda_{c,i}$ and $\mb u_{c,i}$ are the singular values and vectors of $\nabla\mathcal{D}_{\mb\theta}(\mb x_t;\sigma(t),\mb c)$. As shown in~\Cref{fig:guiding effects in nonlinear regime}, this approach fails to improve image quality and frequently produces images with oversaturated colors, indicating that not all of these PCs encode the class-specific features—effective guidance must selectively amplify only those that do.

We note that our heuristic guidance serves as a conceptual approximation and may not always perfectly align with actual CFG behavior; in practice, the actual nonlinear CFG yields more stable and consistent results. Due to the black-box nature of deep networks, fully characterizing this mechanism remains challenging, and we regard this as an important direction for future research.

\vspace{-0.1in}

\section{Discussion and Conclusion}
The experiments in the main-text are conducted extensively using the EDM-1 model~\cite{karras2022elucidating}, which operates directly in pixel space with $64\times64$ resolution. In~\cref{EDM-2 experiments}, we present complementary results on the EDM-2~\cite{karras2024analyzing} latent diffusion model, which generates images at $512\times512$ resolution.

The key insight of this work is that CFG enhances generation quality by amplifying class-specific features while suppressing generic ones. In the linear setting, this effect emerges from the interplay of three guidance components. Different from previous works which mainly focus on analyzing isotropic Gaussian distributions, our study probes the covariance structures of image data, revealing that salient class-specific features emerge from contrast between class covariances.

Although our analysis is based on linear diffusion model (Gaussian data) assumption, the results remain noteworthy since: (i) CFG significantly enhances the generation quality of linear diffusion models, making the linear setting a meaningful stand-alone testbed for studying CFG and (ii) real-world diffusion models can be well-approximated by their linear counterparts for a wide range of noise levels. We note that the dynamics of linear setting is by itself complex: an interpretable solution to the linear reverse ODE is unattainable unless additional assumptions are imposed on the covariance structures (e.g., the common principal components assumption). A natural next step is to extend the analysis to Gaussian mixtures. We have made some initial attempts in~\cref{analyze CFG in Gaussian Mixtures}, showing that CFG guidance in the Gaussian mixture setting can be decomposed in a similar manner as the single Gaussian case.

We believe our findings open several promising directions for future research. First, the observed lack of class-specificity issue implies the current training procedures for diffusion models remain suboptimal. This highlights the need for developing principled training objectives that explicitly encourage the model to learn class-specific patterns. Second, beyond the context of CFG, PCA has been widely utilized for extracting visual features or semantic concepts from diffusion models~\cite{park2023understanding,mo2024freecontrol,gandikota2025sliderspace}. Our results suggest that extending these approaches with Contrastive PCA can be a promising next step for more controllable and interpretable generation. 

\clearpage
\section*{Data Availability Statement}
The code and instructions for reproducing the experiment results will be made available in the following link: \url{https://github.com/Morefre/Towards-Understanding-the-Mechanisms-of-Classifier-Free-Guidance.git}.

\section*{Acknowledgment}
We acknowledge funding support from NSF CCF-2212066, NSF CCF-
2212326, NSF IIS 2402950, and ONR N000142512339. This research used the Delta advanced computing and data resource which is supported by the National Science Foundation (award OAC 2005572) and the State of Illinois. Delta is a joint effort of the University of Illinois Urbana-Champaign and its National Center for Supercomputing Applications\cite{boerner2023access}. We thank Mr. Yixiang Dai for fruitful discussions and valuable feedbacks.


\bibliography{ref}


\appendix
\onecolumn
\begin{center}
{\LARGE \bf Appendices}
\end{center}\vspace{-0.15in}
\par\noindent\rule{\textwidth}{1pt}



\section{Contrastive Principal Component Analysis}
\label{CPC appendix explaination}
Principal component analysis (PCA) finds the features that explain the most variances in the dataset, however, features with high variance do not necessarily correspond to distinct patterns of the target class. 

To discover low-dimensional structure that is unique to a dataset, the work~\cite{abid2018exploring} proposes the contrastive principal component analysis (CPCA), which includes a \emph{background} (or reference) dataset to highlight patterns unique to the \emph{target} dataset. Let $X$ and $Y$ be two datasets with covariance matrices $\mb \Sigma_X$ and $\mb \Sigma_Y$, respectively.
For a unit vector $\mb v\in \mathbb{S}^{d-1}$, its variances $\text{Var}_{X}(\mb v)$ and $\text{Var}_{Y}(\mb v)$ in the two datasets are:
\begin{align}
  \text{Var}_{X}(\mb v) &:= \mb v^T \mb\Sigma_{X} \mb v, 
  & \text{Var}_{Y}(\mb v) &:= \mb v^T \mb\Sigma_{Y} \mb v.
\end{align}
If $\mb v$ corresponds to a unique class-specific pattern of $X$, we expect $\text{Var}_{X}(\mb v)\gg\text{Var}_{Y}(\mb v)$, i.e. it explains significantly more variance in $X$ than in $Y$. Such directions, called the \emph{contrastive principal components (CPCs)}, can be found by iteratively solving:
\begin{align}
    \label{cpc objective appendix}
    \argmax_{\mb v\in \mathbb{S}^{d-1}}\mb v^T(\mb \Sigma_{X}-\mb \Sigma_{Y})\mb v,
\end{align}
 where, at each iteration, the resulting $\mb v$ is subtracted from $\mb \Sigma_{X}-\mb \Sigma_{Y}$. These directions are essentially the eigenvectors of $\mb\Sigma(X)-\mb\Sigma(Y)$. 
Conversely, directions $\mb v$ for which $\text{Var}_{X}(\mb v)\approx\text{Var}_{Y}(\mb v)$ represent either universal structures shared by both $X$ and $Y$ or meaningless features lying in the null space of the data covariances—and are thus discarded as less interesting.

\textbf{Geometric Interpretation of CPCA.} Geometrically, the first $k$ CPCs span the $k$-dimensional subspace that best fits the dataset $X$ while being as far as possible from $Y$~\cite{li2020probabilistic}. This is proved by the following theorem:
 \begin{theorem}
     Without loss of generality, assume $p_X(\mb x)$ and $p_Y(\mb y)$ have zero means (i.e., the data is centered). Then the following objective is equivalent to~\eqref{cpc objective appendix}:
     \begin{align}
         \argmin_{\mb v\in \mathbb{S}^{d-1}}\mathbb{E}_{\mb x\sim p_{X}}||\mb x-\mb v\mb v^T\mb x||_2^2-\mathbb{E}_{\mb y\sim p_{Y}}||\mb y-\mb v\mb v^T\mb y||_2^2.
     \end{align}    
 \end{theorem}
\emph{Proof}:
     \begin{align*}
         \mb v&= \argmin_{\mb v\in \mathbb{S}^{d-1}}\mathbb{E}_{\mb x\sim p_{X}}||\mb x-\mb v\mb v^T\mb x||_2^2-\mathbb{E}_{\mb y\sim p_{Y}}||\mb y-\mb v\mb v^T\mb y||_2^2\\
         &=\argmin_{\mb v\in \mathbb{S}^{d-1}}\mathbb{E}_{\mb x}(\mb x^T\mb x-\mb x^T\mb v\mb v^T\mb x)-\mathbb{E}_{\mb y}(\mb y^T\mb y-\mb y^T\mb v\mb v^T\mb y)\\
         &=\argmin_{\mb v\in \mathbb{S}^{d-1}}\mathbb{E}_{\mb x}(-\mb x^T\mb v\mb v^T\mb x)-\mathbb{E}_{\mb y}(-\mb y^T\mb v\mb v^T\mb y)\\
         &=\argmin_{\mb v\in \mathbb{S}^{d-1}}-\mathbb{E}_{\mb x}(\mb x^T\mb v\mb v^T\mb x)+\mathbb{E}_{\mb y}(\mb y^T\mb v\mb v^T\mb y)\\
         &=\argmax_{\mb v\in \mathbb{S}^{d-1}}\mathbb{E}_{\mb x}(\mb x^T\mb v\mb v^T\mb x)-\mathbb{E}_{\mb y}(\mb y^T\mb v\mb v^T\mb y)\\
         &= \argmax_{\mb v\in \mathbb{S}^{d-1}}\mb v^T\mathbb{E}_{\mb x}(\mb x\mb x^T)\mb v-\mb v^T\mathbb{E}_{\mb y}(\mb y\mb y^T)\mb v\\
         &=\argmax_{\mb v\in \mathbb{S}^{d-1}}\mb v^T(\mb \Sigma_{X}-\mb \Sigma_{Y})\mb v.
     \end{align*}
 This proof is adapted from~\cite{li2020probabilistic}.

\section{Analytical Solution to the Reverse Diffusion ODE} 
\label{sec:proof}
In this section, we examine the solutions to both the naive reverse diffusion ODE~\eqref{reverse ode} and the CFG-guided reverse diffusion ODE~\eqref{cfg sampling} in the context of linear diffusion models. We follow the EDM formulation~\cite{karras2022elucidating}, which uses time schedule $\sigma(t)=t$. Notice that this same schedule is also used by the well-known DDIM sampler~\cite{songdenoising}.

\subsection{Naive Diffusion Reverse ODE}
\label{naive linear diffusion ode}
We begin by analyzing the diffusion ODE~\eqref{reverse ode} with no guidance applied. The proof below is borrowed from~\cite{wangunreasonable}. 

Let $\mb\mu$ and $\mb\Sigma$ be the mean and covariance of $p_\text{data}(\mb x)$ respectively. Suppose $\mb U\mb\Lambda\mb U^T$ is the full SVD of $\mb \Sigma$ with $\mb U\in \mathbb{R}^{d\times d}$ being orthonormal and $\mb \Lambda=\text{diag}(\lambda_1,\cdots,\lambda_d)$ contains the singular values. For image datasets, the covariance is often low-rank implying some singular values are $0$. Under the constraint that $\mathcal{D}(\mb x;\sigma (t))$ is linear (with a bias term), the optimal solution to~\eqref{denoising score matching} has the closed-form~\cite{liunderstanding}:
\begin{align}
    \label{linear Gaussian diffusion model 1}
    \mathcal{D}(\mb x;\sigma (t)) = \mb \mu+\mb U\Tilde{\mb \Lambda}_{\sigma(t)}\mb U^T (\mb x-\mb \mu),
\end{align}
where $\Tilde{\mb \Lambda}_{\sigma(t)}=\diag \paren{\frac{\lambda_1}{\lambda_1+\sigma^2(t)}, \cdots,
     \frac{\lambda_d}{\lambda_d+\sigma^2(t)} }$. This optimal linear solution is obtained by setting the derivative of \eqref{denoising score matching} with respect to the weight and bias to zero, leveraging the fact that the objective is convex under the linear constraint.
    
     Following the EDM framework, this optimal linear denoiser yields the sampling trajectory for~\eqref{reverse ode} as:
\begin{align}
    d\mb x&=-\sigma\nabla_{\mb x}\log p(\mb x;\sigma)d\sigma\\
   \Rightarrow d\mb x &=\frac{(\mb I-\mb U\tilde{\mb\Lambda}_{\sigma}\mb U^T)(\mb x-\mb \mu)}{\sigma}d\sigma \\
   \Rightarrow d(\mb x-\mb\mu) &=\frac{\mb U(\mb I-\tilde{\mb\Lambda}_{\sigma})\mb U^T(\mb x-\mb \mu)}{\sigma}d\sigma,
\end{align} 
where we omit the subscript $t$ for simplicity.

Define $c_k(\sigma)=\mb u_k^T(\mb x-\mb \mu)$ for $k\in \{1,...,d\}$, we have:
\begin{align}
    dc_k(\sigma)&= \frac{\sigma}{\lambda_k+\sigma^2}c_k(\sigma)d\sigma \\
    \Rightarrow \frac{dc_k(\sigma)}{c_k(\sigma)}&=\frac{\sigma}{\lambda_k+\sigma^2}d\sigma.
    \label{ode 1}
\end{align}
Integrating both sides of~\eqref{ode 1}, we get:
\begin{align}
    d\log c_k(\sigma) &= d\log \sqrt{\lambda_k+\sigma^2} \\
    \Rightarrow c_k(\sigma) &= \sqrt{\lambda_k+\sigma^2}C,
\end{align}
where $C$ is the integral constant. Using the initial condition $c_k(\sigma_T)=\mb u_k^T(\mb x_T-\mb \mu)$, we have:
\begin{align}
    C &= \frac{\mb u_k^T(\mb x_T-\mb \mu)}{\sqrt{\lambda_k+\sigma_T^2}}\\\Rightarrow c_k(\sigma)&=\sqrt{\frac{\lambda_k+\sigma^2}{\lambda_k+\sigma_T^2}}\mb u_k^T(\mb x_T-\mb \mu) \\
    \Rightarrow \mb x_t &= \mb\mu + \sum_{k=1}^d\sqrt{\frac{\lambda_k+\sigma_t^2}{\lambda_k+\sigma_T^2}}\mb u_k^T(\mb x_T-\mb \mu)\mb u_k,
    \label{no guidance solution}
\end{align}
where the last equality holds because $\mb x_t-\mb\mu=\sum_{i=k}^dc_k(\sigma_t)\mb u_k$

Notice that the generated samples are primarily determined by the data’s covariance structure. However, since the covariance may not capture the most distinctive features of a specific class, the resulting images often lack sufficient class specificity.

\subsection{CFG-Guided ODE}
\label{CFG guidaed ode proof}
To apply CFG, we need two separate models corresponding to conditional and unconditional data respectively. Let $\mb\mu_c$, $\mb\mu_{uc}$ be the means of the conditional and unconditional data, and let $\mb\Sigma_c=\mb U_c\mb\Lambda_c\mb U_c^T$, $\mb\Sigma_{uc}=\mb U_{uc}\mb\Lambda_{uc}\mb U_{uc}^T$ be their corresponding covariances, where $\mb \Lambda_c=\text{diag}(\lambda_{c,1},\cdots,\lambda_{c,d})$ and $\mb \Lambda_{uc}=\text{diag}(\lambda_{uc,1},\cdots,\lambda_{uc,d})$. Then the conditional and unconditional optimal linear denoisers take the following forms:
\begin{align}
    \label{conditional optimal linear denoiser}
    \mathcal{D}_{\mathrm{L}}(\mb x;\sigma (t),\mb c) = \mb \mu_c+\mb U_c\Tilde{\mb \Lambda}_{c,\sigma(t)}\mb U_c^T (\mb x-\mb \mu_c),\\
    \label{unconditional optimal linear denoiser}
    \mathcal{D}_{\mathrm{L}}(\mb x;\sigma (t)) = \mb \mu_{uc}+\mb U_{uc}\Tilde{\mb \Lambda}_{uc,\sigma(t)}\mb U_{uc}^T (\mb x-\mb \mu_{uc}),
\end{align}

Then the CFG sampling trajectory~\eqref{cfg sampling} can be expressed in terms of the optimal linear denoisers:
\begin{align}
    d\mb x_t &= -\sigma(t)(\frac{\mathcal{D}_{\mathrm{L}}(\mb x_t;\sigma(t),\mb c)-\mb x_t}{\sigma^2(t)}+\gamma\frac{\mathcal{D}_{\mathrm{L}}(\mb x_t;\sigma(t),\mb c)-\mathcal{D}_{\mathrm{L}}(\mb x_t;\sigma(t))}{\sigma^2(t)})dt \label{term all}\\
    &=-\frac{1}{\sigma(t)}(\mb U_c\Tilde{\mb\Lambda}_{\sigma(t),c}\mb U_c^T-\mb I)(\mb x_t-\mb\mu_c) \label{term 1}dt\\
    &\quad-\frac{\gamma}{\sigma(t)}(\mb U_c\Tilde{\mb\Lambda}_{\sigma(t),c}\mb U_c^T-\mb U_{uc}\Tilde{\mb\Lambda}_{\sigma(t),uc}\mb U_{uc}^T)(\mb x_t-\mb\mu_c)dt \label{term 2}\\
    &\quad-\frac{\gamma}{\sigma(t)}(\mb I-\mb U_{uc}\Tilde{\mb\Lambda}_{\sigma(t),uc}\mb U_{uc}^T)(\mb \mu_c-\mb \mu_{uc})dt\label{term 3},
\end{align}
where~\eqref{term 1} is the naive conditional score while~\eqref{term 2} and~\eqref{term 3} together form the CFG guidance direction. Note that under the setting of linear diffusion model, we have 
\begin{align}
p(\mb x;\sigma(t)) &= \mathcal{N}(\mb\mu_{uc}, \mb\Sigma_{uc}+\sigma^2(t)\mb I),\\
p(\mb x|\mb c;\sigma(t)) &= \mathcal{N}(\mb\mu_{c}, \mb\Sigma_{c}+\sigma^2(t)\mb I),\\
\nabla_{\mb x}\log p(\mb x;\sigma(t))&=(\mb\Sigma_{uc}+\sigma^2(t)\mb I)^{-1}(\mb\mu_{uc}-\mb x),\\
\nabla_{\mb x}\log p(\mb x|\mb c;\sigma(t))&=(\mb\Sigma_{c}+\sigma^2(t)\mb I)^{-1}(\mb\mu_{c}-\mb x),\\
\mathbb{E}_{\mb x\sim p(\mb x|\mb c;\sigma(t))}[\nabla_{\mb x}\log p(\mb x|\mb c;\sigma(t))-\nabla_{\mb x}\log p(\mb x;\sigma(t))]&=(\mb\Sigma_{uc}+\sigma^2(t)\mb I)^{-1}(\mb\mu_{c}-\mb\mu_{uc})\\
&=\frac{1}{\sigma^2(t)}(\mb I-\Tilde{\mb \Sigma}_{uc,t} )(\mb \mu_c-\mb \mu_{uc}).
\end{align}
Therefore, we have:
\begin{align}
    g_{mean}(t)&=\gamma\mathbb{E}_{\mb x\sim p(\mb x|\mb c;\sigma(t))}[\nabla_{\mb x}\log p(\mb x|\mb c;\sigma(t))-\nabla_{\mb x}\log p(\mb x;\sigma(t))]\\
    &=\gamma\mathbb{E}_{\mb x\sim p(\mb x|\mb c;\sigma(t))}[\nabla_{\mb x}\log\frac{p(\mb x|\mb c;\sigma(t))}{p(\mb x;\sigma(t))}]
\end{align}
which implies the mean-shift guidance term can be interpreted as the probability-weighted average of the steepest ascent direction that maximizes the log-likelihood ratio of the noise mollified conditional and unconditional distributions.

Since~\eqref{term all} is a first-order non-homogeneous differential equation, its closed-form solution can in principle, be expressed through integrals. However, these integrals cannot be explicitly evaluated or decoupled in the general case. To obtain a tractable, interpretable solution, we must impose additional assumptions on the structures of $\mb\Sigma_c$ and $\mb\Sigma_{uc}$. Therefore, we make the following assumptions:

\paragraph{Assumption.} \textit{The covariance matrices $\mb\Sigma_c$ and $\mb\Sigma_{uc}$ are simultaneously diagonalizable, i.e., $\mb\Sigma_\text{uc}=\mb U_c\mb \Lambda_{uc}\mb U_c^T$, where $\mb U_c\in \mathbb{R}^d$ are the principal components (singular vectors) of the conditional data. Here $\mb\Lambda_{uc}$ is not necessarily ordered by the magnitude of the singular value.}

This is well-known as the \textit{Common Principal Components assumption}~\cite{flury1984common,10.5555/74760}, widely utilized to analyze structural relationships across data groups. Under this assumption, the relative importance of the $k^\text{th}$ principal component $\mb u_{c,k}$ in the conditional and unconditional datasets is fully determined by the relative magnitudes of its associated singular values:
\begin{itemize}[leftmargin=*]
    \item If $\lambda_{c,k}>\lambda_{uc,k}$, then $\mb u_{c,k}$ explains more variance in the conditional dataset than in the unconditional dataset, i.e., it is more distinct for the conditional data. This corresponds to the positive CPC discussed in the main text.

    \item Conversely, if $\lambda_{c,k}<\lambda_{uc,k}$, then $\mb u_{c,k}$ explains more variance in the unconditional dataset than in the conditional dataset, making it more distinct for the unconditional data and therefore it is a negative CPC.
\end{itemize}


Under the assumption, the CFG guided ODE~\eqref{term all} can be simplified as:
\begin{align}
    d\mb x = \frac{(\mb I-\mb U_c\tilde{\mb\Lambda}_{\sigma,c}\mb U_c^T)(\mb x-\mb\mu_c)}{\sigma}d\sigma&-\gamma\frac{\mb U_c(\tilde{\mb \Lambda}_{\sigma,c} -\tilde{\mb \Lambda}_{\sigma,uc})\mb U_c^T(\mb x-\mb\mu_c)}{\sigma}d\sigma\\
    &-\gamma \frac{\mb U_c(\mb I-\tilde{\mb\Lambda}_{\sigma,uc})\mb U_c^T (\mb\mu_c-\mb\mu_{uc})}{\sigma}d\sigma.
\end{align}
Define $c_k(\sigma)=\mb u_{c,k}^T(\mb x-\mb \mu_c)$, we have:
\begin{align}
    dc_k(\sigma)=\frac{\sigma}{\lambda_{c,k}+\sigma^2}c_k(\sigma)d\sigma&-\gamma\frac{\sigma(\lambda_{c,k}-\lambda_{uc,k})}{(\lambda_{c,k}+\sigma^2)(\lambda_{uc,k}+\sigma^2)}c_k(\sigma)d\sigma \\
    &-\gamma\frac{\sigma}{\lambda_{uc,k}+\sigma^2}\mb u_{c,k}^T(\mb\mu_c-\mb\mu_{uc})d\sigma.
    \label{first-order linear ode}
\end{align}
Therefore, the dynamics of $dc_k(\sigma)$ can be expressed as:
\begin{align}
    dc_k(\sigma)+f(\sigma)c_k(\sigma)d\sigma =g(\sigma)d\sigma,
    \label{simplified homogeneous ODE}
\end{align},
where $f(\sigma)=-(\frac{\sigma}{\lambda_{c,k}+\sigma^2}-\gamma\frac{\sigma(\lambda_{c,k}-\lambda_{uc,k})}{(\lambda_{c,k}+\sigma^2)(\lambda_{uc,k}+\sigma^2)})$ and $g(\sigma)=-\gamma\frac{\sigma}{\lambda_{uc,k}+\sigma^2}\mb u_{c,k}^T(\mb\mu_c-\mb\mu_{uc})$. 
\paragraph{Homogeneous ODE.} We first consider the homogeneous counterpart of~\eqref{simplified homogeneous ODE}:
\begin{align}
    dc_k(\sigma) = -f(\sigma)c_k(\sigma)d\sigma,
    \label{homogeneous part}
\end{align}
where $-f(\sigma)c_k(\sigma)d\sigma$ corresponds to the combination of the standard conditional score~\eqref{term 1} and the CPC guidance term~\eqref{term 2}. Integrating over both sides of~\eqref{homogeneous part}, we get:
\begin{align}
    c_k(\sigma)=Ce^{\int -f(\sigma)d\sigma}.
\end{align}
Notice that:
\begin{align}
    \int -f(\sigma) d\sigma&=\int \frac{\sigma}{\lambda_{c,k}+\sigma^2}d\sigma-\gamma\int \frac{\sigma(\lambda_{c,k}-\lambda_{uc,k})}{(\lambda_{c,k}+\sigma^2)(\lambda_{uc,k}+\sigma^2)}d\sigma \\
    &= \frac{1}{2}\ln (\lambda_{c,k}+\sigma^2)+\frac{\gamma}{2}\ln (\frac{\lambda_{c,k}+\sigma^2}{\lambda_{uc,k}+\sigma^2}),
\end{align}
which implies:
\begin{align}
    c_k(\sigma)=C(\lambda_{c,k}+\sigma^2)^{\frac{1}{2}}(\frac{\lambda_{c,k}+\sigma^2}{\lambda_{uc,k}+\sigma^2})^{\frac{\gamma}{2}}.
\end{align}
Applying the initial condition that $c_k(\sigma_T)=\mb u_{c,k}^T(\mb x_T-\mb \mu_c)$, we have:
\begin{align}
    C&=(\lambda_{c,k}+\sigma_T^2)^{-\frac{1}{2}}(\frac{\lambda_{c,k}+\sigma_T^2}{\lambda_{uc,k}+\sigma_T^2})^{-\frac{\gamma}{2}} \mb u_{c,k}^T(\mb x_T-\mb \mu_c)\\
    \\
    \Rightarrow c_k(\sigma_t)&=\sum_{k=1}^d(\frac{\lambda_{c,k}+\sigma_t^2}{\lambda_{c,k}+\sigma_T^2}\frac{\lambda_{uc,k}+\sigma_T^2}{\lambda_{uc,k}+\sigma_t^2})^{\frac{\gamma}{2}}\sqrt{\frac{\lambda_{c,k}+\sigma_t^2}{\lambda_{c,k}+\sigma_T^2}}\mb u_{c,k}^T(\mb x_T-\mb \mu_c) \\
    \Rightarrow \mb x_t&= \mb\mu_c+\sum_{k=1}^dh(\lambda_{c,k},\lambda_{uc,k})^\frac{\gamma}{2} \sqrt{\frac{\lambda_{c,k}+\sigma_t^2}{\lambda_{c,k}+\sigma_T^2}}\mb u_{c,k}^T(\mb x_T-\mb \mu_c)\mb u_{c,k},
\end{align}
where $h(\lambda_{c,k},\lambda_{uc,k})=\frac{\lambda_{c,k}+\sigma(t)^2}{\lambda_{c,k}+\sigma^2(T)}\frac{\lambda_{uc,k}+\sigma^2(T)}{\lambda_{uc,k}+\sigma^2(t)}$.
Compared with the solution to the naive reverse process with no guidance~\eqref{no guidance solution}, each component of $\mb x_t$ differs only by a scalar factor $h(\lambda_{c,k},\lambda_{uc,k})^\frac{\gamma}{2}$. Specifically:
\begin{itemize}[leftmargin=*]
    \item $h(\lambda_{c,k},\lambda_{uc,k})\geq 1$ if and only if $\lambda_{c,k}\geq\lambda_{uc,k}$, meaning positive CPCs are enhanced (scaled up).
    \item $h(\lambda_{c,k},\lambda_{uc,k})\leq 1$ if and only if $\lambda_{c,k}\leq\lambda_{uc,k}$, meaning negative CPCs are suppressed (scaled down).
\end{itemize}
Note that the guidance strength $\gamma$ provides additional control, amplifying or reducing the degree of enhancement or suppression for each component. 

\paragraph{Non-Homogeneous ODE.} Let $\hat{c}_k(\sigma)$ be the solution to the homogeneous ODE~\eqref{homogeneous part}, then the solution to the non-homogeneous ODE~\eqref{simplified homogeneous ODE} takes the form:
\begin{align}
    c_k(\sigma) = \hat{c}_k(\sigma)  + \frac{1}{I(\sigma)}\int I(\sigma')g(\sigma')d\sigma',
\end{align}
where $I(\sigma)=e^{\int f(\sigma')d\sigma'}$ is the integrating factor. Since:
\begin{align}
    I(\sigma) = C(\lambda_{c,k}+\sigma^2)^{-\frac{1}{2}}(\frac{\lambda_{c,k}+\sigma^2}{\lambda_{uc,k}+\sigma^2})^{-\frac{\gamma}{2}},
\end{align}
we have:
\begin{align}
    c_k(\sigma) &= \hat{c}_k(\sigma) + \gamma(\lambda_{c,k}+\sigma^2)^{\frac{1}{2}}(\frac{\lambda_{c,k}+\sigma^2}{\lambda_{uc,k}+\sigma^2})^{\frac{\gamma}{2}}\int_{\sigma}^{\sigma_T}\frac{(\lambda_{uc,k}+\tilde{\sigma}^2)^{\frac{\gamma}{2}-1}}{(\lambda_{c,k}+\tilde{\sigma}^2)^{\frac{\gamma+1}{2}}}\mb u_{c,k}^T(\mb\mu_c-\mb\mu_{uc})\tilde{\sigma} d\tilde{\sigma}\\
    &= \hat{c}_k(\sigma) + \gamma b_{\sigma, k}\mb u_{c,k}^T(\mb\mu_c-\mb\mu_{uc}),
\end{align}
where $b_{\sigma,k}=(\lambda_{c,k}+\sigma^2)^{\frac{1}{2}}(\frac{\lambda_{c,k}+\sigma^2}{\lambda_{uc,k}+\sigma^2})^{\frac{\gamma}{2}}\int_{\sigma}^{\sigma_T}\frac{(\lambda_{uc,k}+\tilde{\sigma}^2)^{\frac{\gamma}{2}-1}}{(\lambda_{c,k}+\tilde{\sigma}^2)^{\frac{\gamma+1}{2}}}\tilde{\sigma} d\tilde{\sigma}$. Therefore we have:
\begin{align}
    \mb x_t&= \mb\mu_c+\sum_{k=1}^dh(\lambda_{c,k},\lambda_{uc,k})^\frac{\gamma}{2} \sqrt{\frac{\lambda_{c,k}+\sigma_t^2}{\lambda_{c,k}+\sigma_T^2}}\mb u_{c,k}^T(\mb x_T-\mb \mu_c)\mb u_{c,k}+\gamma\sum_{k=1}^d b_k\mb u_{c,k} \mb u_{c,k}^T(\mb\mu_c-\mb\mu_{uc}) \\
    &=  \mb\mu_c+\sum_{k=1}^dh(\lambda_{c,k},\lambda_{uc,k})^\frac{\gamma}{2} \sqrt{\frac{\lambda_{c,k}+\sigma_t^2}{\lambda_{c,k}+\sigma_T^2}}\mb u_{c,k}^T(\mb x_T-\mb \mu_c) \mb u_{c,k}+ \gamma\mb U_c\mb B_{\sigma_t}\mb U_c^T(\mb\mu_c-\mb\mu_{uc}), 
\end{align}
where $\mb B_{\sigma_t}=\text{diag}(b_{\sigma_t,1},...,b_{\sigma_t,d})$. Here $b_k$ depends only on $\lambda_{uc,k}, \lambda_{c,k}$ and $\sigma(t)$. Hence the mean-shift guidance term~\eqref{term 3} has the effect of adding constant perturbations that are independent of the initial noise $\mb x_T$ to the sampling trajectory.
\subsection{Empirical Verification on Synthetic Data.}
\label{subsec: empirical verification}
\vspace{-0.1in}
\begin{figure*}[t!]
    \centering
    \includegraphics[width=0.9\linewidth]{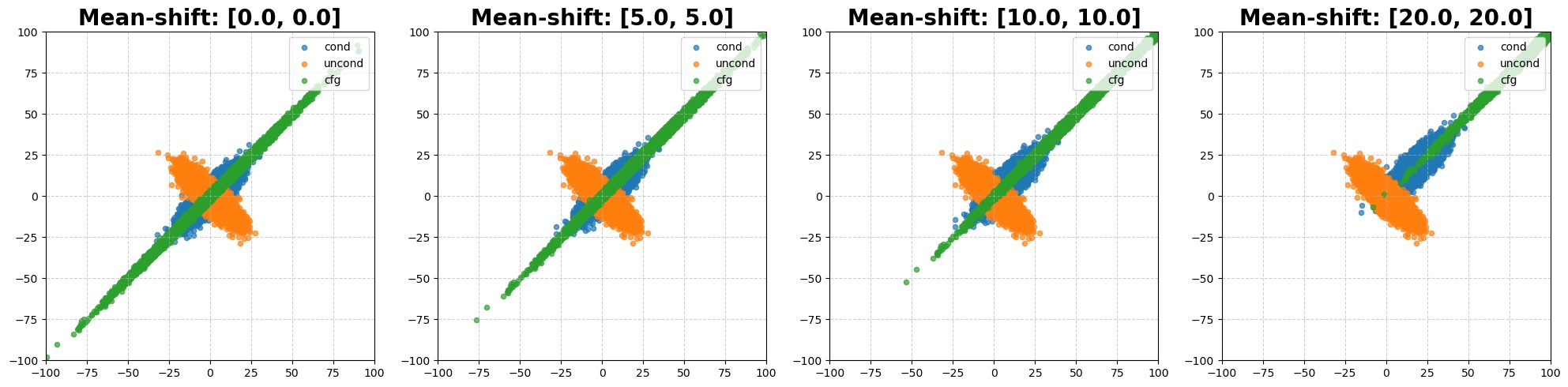}
  \caption{\textbf{CFG effects in 2D.} Each subplot differs by the class mean \(\boldsymbol\mu_c\), indicated in the titles. Blue, orange and green points show 1{,}000 samples generated from conditional sampling,  naive unconditional sampling and CFG sampling, respectively.}
  \label{2d-cfg}
\end{figure*}

We validate~\Cref{theorem} on a 2D toy model with
$\mb U_c = \left[
\begin{smallmatrix}
\frac{1}{\sqrt{2}} & \frac{1}{\sqrt{2}} \\
\frac{1}{\sqrt{2}} & \frac{-1}{\sqrt{2}}
\end{smallmatrix}\right]$, 
$\mb\Lambda_c=\left[
\begin{smallmatrix}
10 & 0 \\
0 & 3 
\end{smallmatrix}\right]$ and
$\mb\Lambda_{uc}=\left[
\begin{smallmatrix}
3 & 0 \\
0 & 10 
\end{smallmatrix}\right]$. 
\Cref{2d-cfg} shows the effects of CFG under different class mean $\mb{\mu}_c$ (with $\gamma = 1$ and $\mb{\mu}_{uc} = \mb{0}$). As predicted, CFG enhances variation along the positive CPC $\left[\begin{smallmatrix}\frac{1}{\sqrt{2}} & \frac{1}{\sqrt{2}}\end{smallmatrix}\right]^T$, suppresses variation along the negative CPC $\left[\begin{smallmatrix}\frac{1}{\sqrt{2}} & -\frac{1}{\sqrt{2}}\end{smallmatrix}\right]^T$, and shifts samples roughly toward $\mb\mu_c-\mb\mu_{uc}$ at a rate proportional to $\gamma ||\mb{\mu}_c-\mb{\mu}_{uc}||_2$.

\section{Constructing Linear Denoisers}
Constructing the linear denoisers~\eqref{linear Gaussian diffusion model} requires estimating the data means and covariances. We perform our experiments on CIFAR-10~\cite{krizhevsky2009learning} and ImageNet dataset~\cite{deng2009imagenet}, estimating the linear denoisers for each dataset in different ways:
\begin{itemize}[leftmargin=*]
    \item \textbf{CIFAR-10.} This dataset consists of 10 different classes, each with 5000 images. We obtain the unconditional linear diffusion model by computing the empirical mean and covariance across all 50000 images. For conditional linear diffusion model, we construct a separate linear model for each class, using that class's mean and covariance estimated from the 5000 images.
    \item \textbf{ImageNet.} This dataset contains 1000 classes, each with approximately 1000 images—a smaller per-class sample size that can introduce bias when estimating means and covariances directly from the training set. Although such direct estimation still yields linear denoisers aligned with the actual diffusion models in the linear regime, these denoisers tend to generate noisier images. We hypothesize that, in the conditional setting, each class’s diffusion denoiser may implicitly leverage information from other classes, meaning the true mean and covariance learned by the deep diffusion model can differ (albeit slightly) from the those estimated solely from that class’s data. To obtain a more accurate linear approximation, we therefore generate 50,000 samples per class with the trained diffusion model, then compute the empirical mean and covariance from these generated samples. Nonetheless, all of our main conclusions remain valid even if we build the linear models using the actual ImageNet training data.
\end{itemize}

\section{Naive Conditional Generation Lacks Class-Specificity}
\label{sec: more discussion on covariance structure}
In~\cref{naive conditional generation is sub-optimal} we argue that naive conditional generation lacks class-specificity and in the linear model setting, such issue can be partially attributed to the non-distinctiveness of the class covariance matrices. In this section, we provide comprehensive experiments to support our claim both qualitatively and quantitatively.

\subsection{Qualitative Results} 
\label{subsec: Qualitative Results}

\begin{figure}[t!]
    \centering
    \includegraphics[width=1\linewidth]{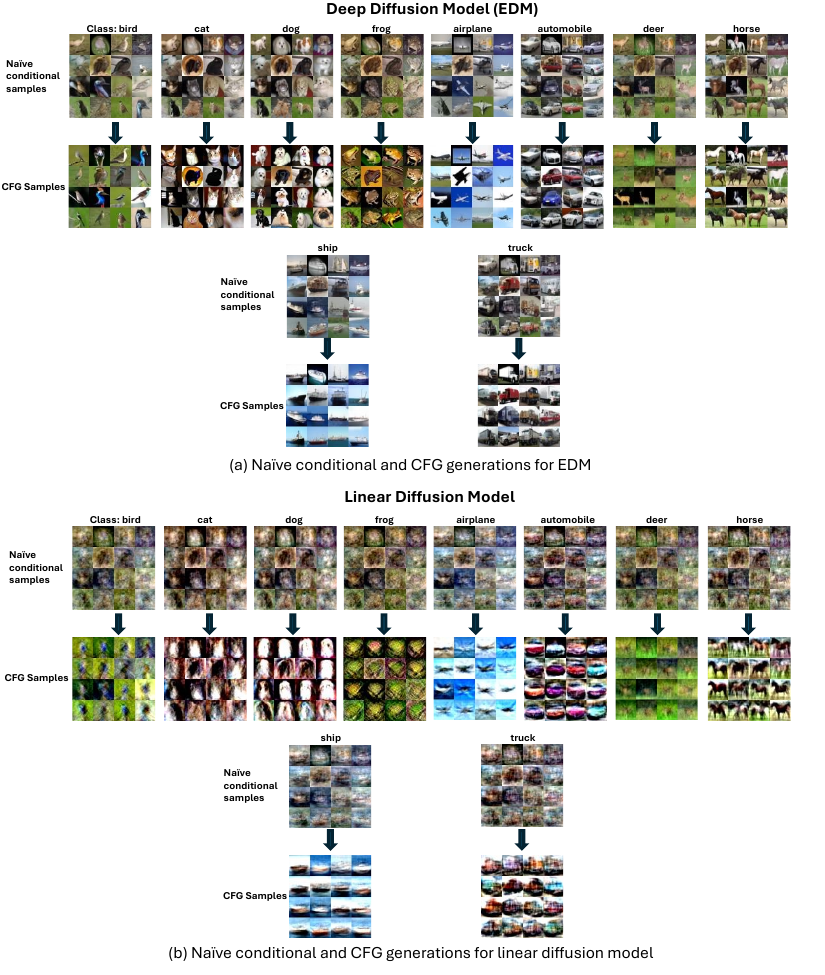}
    \caption{\textbf{Effects of CFG on CIFAR-10.} (a) and (b) demonstrate the naive conditional samples and the CFG-guided samples of deep diffusion model and linear diffusion model respectively. Each grid corresponds to the same initial noise.}
    \label{fig:cifar-10 conditonal vs cfg linear and nonlinear}
\end{figure}

\begin{figure}[t!]
    \centering
    \includegraphics[width=1\linewidth]{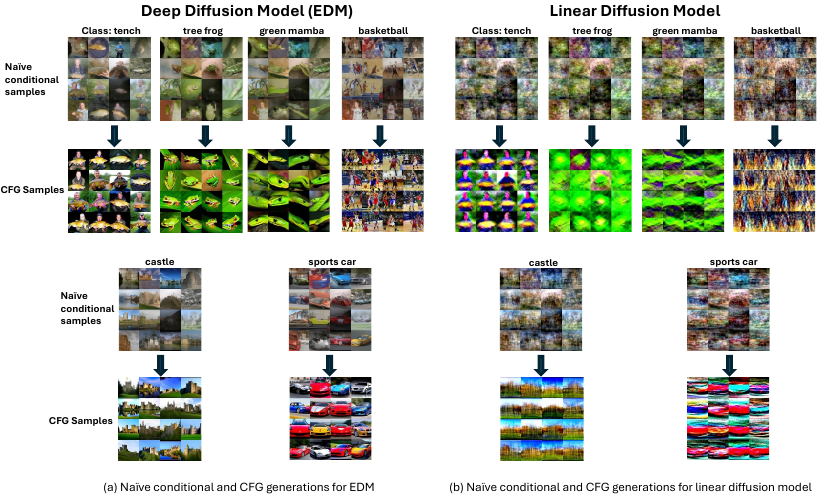}
    \caption{\textbf{Effects of CFG on ImageNet.} (a) and (b) demonstrate the naive conditional samples and the CFG-guided samples of deep diffusion model and linear diffusion model respectively. Each grid corresponds to the same initial noise. Here we only display 6 classes since the other 4 classes are presented in~\cref{fig:Gaussian inductive bias}.}
    \label{fig:ImageNet conditonal vs cfg linear and nonlinear}
\end{figure}

We generate samples using naive conditional sampling~\eqref{conditional sampling} and CFG sampling~\eqref{cfg sampling} for all 10 classes of CIFAR-10, as well as for 10 selected ImageNet classes: including (i) class 0: tench , (ii) class 31: tree frog, (iii) class 64: green mamba, (iv) class 207: golden retriever, (v) class 430: basketball, (vi) class 483: castle, (vii) class 504: coffee mug, (viii) class 817: sports car, (ix) class 933: cheese burger and (x) class 947: mushroom. CFG is applied to the entire noise interval $\sigma(t)\in[0.002,80]$, with guidance strength $\gamma=4$. The results for CIFAR-10 and ImageNet are shown in~\Cref{fig:cifar-10 conditonal vs cfg linear and nonlinear} and~\Cref{fig:ImageNet conditonal vs cfg linear and nonlinear} respectively.

Our key observations are as follows:
\begin{itemize}[leftmargin=*]
    \item \textbf{Linear Diffusion Models.} Despite being built from class-specific means and covariances, the conditional \emph{linear} diffusion models produce visually similar samples that lack distinguishable class features. From \eqref{Gaussian ODE}, we see that the generated samples are largely shaped by each class’s covariance structure; hence, their indistinct and low-quality generations suggest that these covariance matrices are insufficiently distinctive.
    
    \item \textbf{Deep diffusion models (EDM)} These models inherit similar limitations. The generated samples often exhibit poor image quality, with incoherent features that blend into the background and the class-specific image structures can be hard to discern. Furthermore, images generated from the same initial noise can appear structurally similar even under different class labels, indicating that naive conditional sampling fails to capture distinct, class-specific patterns. Lastly, comparing the generations from linear model and EDM reveals they match in terms of the overall structures, underscoring the key role of covariance in shaping higher-level features. Consequently, when class-specific covariance matrices are not sufficiently distinct, sample quality remains limited—even in nonlinear models.
\end{itemize}

\subsection{Quantitative Results}
\label{subsec: Quantitative Results}

To quantify the class-specificity gap, we compare the pairwise class similarity with FID score~\cite{heusel2017gans}, which measures the similarity between two datasets $X$ and $Y$ in the Inception embedding space. For every ordered pair of different classes $(c_i, c_j)$ we build two datasets $(X, Y)$ and compute $\text{FID}(X,Y)$ under three settings:
\begin{itemize}[leftmargin=*]
  \item Real data.  
        $X$ and $Y$ contain all training images from classes $c_i$ and $c_j$, respectively.
  \item Naive conditional sampling.
        $X$ and $Y$ contain images generated by vanilla conditional sampling \eqref{conditional sampling} with the EDM model. We generate approximately the same number of images as the corresponding training images.
  \item Classifier‑free guidance (CFG). 
        $X$ and $Y$ contain images generated from the same EDM model using CFG sampling \eqref{cfg sampling}. We generate approximately the same number of images as the corresponding training images.
\end{itemize}

\begin{figure}
    \centering
    \includegraphics[width=1\linewidth]{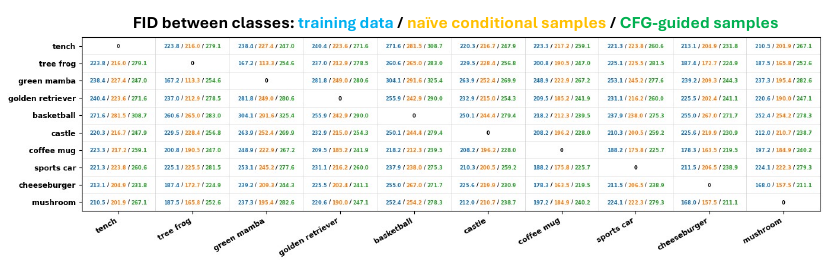}
    \caption{\textbf{Class-to-Class Similarity (Measured with FID)}. Each cell reports the FID between datasets of two classes, built with (i) training data (ii) data generated by naive conditional sampling and (iii) data generated by CFG sampling.}
    \label{fig:class-to-class similarity}
\end{figure}

The results are presented in~\Cref{fig:class-to-class similarity}, which shows that for most pairs of classes, when $X$ and $Y$ are built with images generated with naive conditional sampling, the FID (colored in orange) is consistently lower than when they are built with training data (colored in blue). Because lower FID indicates higher similarity, this results confirms that images produced by naive conditional sampling are less distinguishable across classes than the real data. In contrast, CFG greatly improves the FID score, implying an increased inter-class separation.

\begin{figure}
    \centering
    \includegraphics[width=1\linewidth]{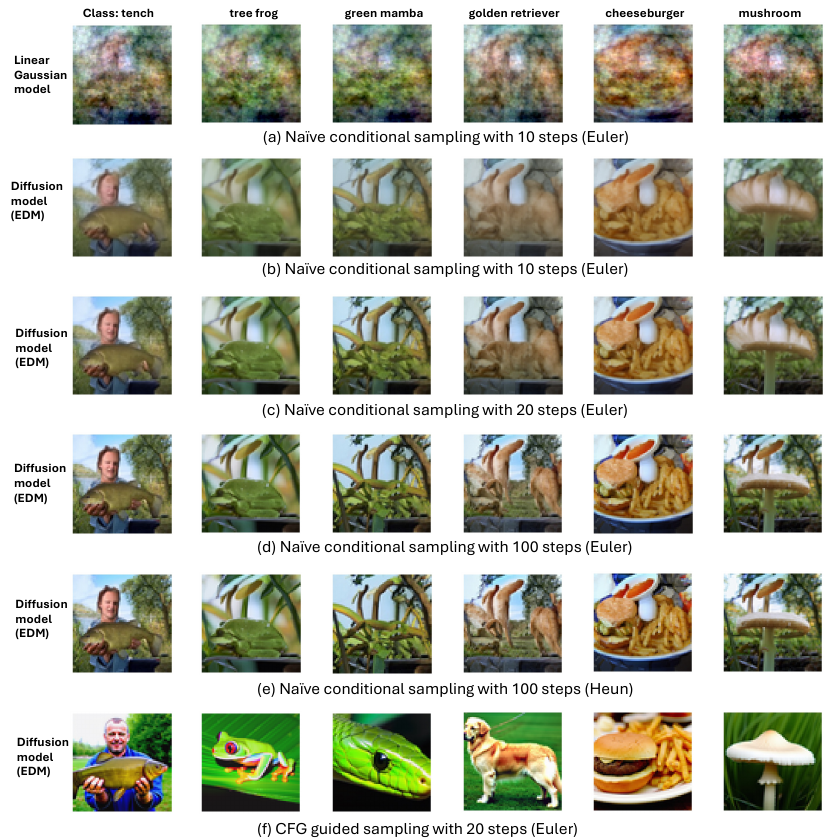}
    \caption{\textbf{Naive conditional sampling lacks class-specific features.} Figure (a) shows the samples generated with naive conditional sampling using linear diffusion models. Figures (b)-(e) show the samples generated with naive conditional sampling using the actual diffusion models with different steps and samplers. Figure (f) shows the generated samples with CFG guided sampling. Note that the generated images from linear models of different classes share high visually similarity, implying the covariance structures of different classes are not distinctive enough. Similar structural similarity can be observed in the samples of nonlinear diffusion models. CFG greatly alleviates this issue of lack of class-specificity, leading to images with clear class-specific features.}
    \label{fig:various_steps_samples_visualization}
\end{figure}

The samples used for calculating FID in~\Cref{fig:class-to-class similarity} are generated using 20 steps of Euler method (first-order sampler). Increasing the number of steps or switching to higher-order sampler only marginally narrows the gap.~\Cref{tab:avg-similarity-fid} shows the inter-class FID averaged over 10 selected classes as described in~\cref{subsec: Qualitative Results} with different sampling steps and sampler. Note that even when using 100 steps and second-order Heun samplers, the average inter-class FID is still considerably smaller compared to the training data (ground truth).~\Cref{fig:various_steps_samples_visualization} qualitatively visualizes the samples generated from the same initial noise but different class labels. Despite conditioned on different labels, the generated images share high structural similarity. For certain classes, such as tree frog, green mamba and golden retriever, the class features are even hard to discern. In contrast, CFG greatly reduces the structural similarity, yielding images with clear, class-specific features.

\begin{table}[h]
  \centering
  \caption{Average inter‑class FID for training data and various sampling settings (10‑class average).}
  \vspace{\baselineskip}  
  \label{tab:avg-similarity-fid}

  \begin{tabular}{@{}lccc@{}}
    \toprule
    \textbf{Method} & \textbf{Steps} & \textbf{Sampler} & \textbf{Avg.\ FID} \\ \midrule
    Training (ground truth)     & --   & --    & 226.6 \\
    Naive conditional           & 10   & Euler & 210.7 \\
                                & 20   & Euler & 214.6 \\
                                & 30   & Euler & 215.8 \\
                                & 50   & Euler & 215.9 \\
                                & 100  & Euler & 216.2 \\
                                & 100  & Heun  & 216.3 \\ \midrule
    CFG guided ($\gamma=4$)                 & 20   & Euler & 258.9 \\ \bottomrule
  \end{tabular}
\end{table}

\begin{figure}
    \centering
    \includegraphics[width=0.8\linewidth]{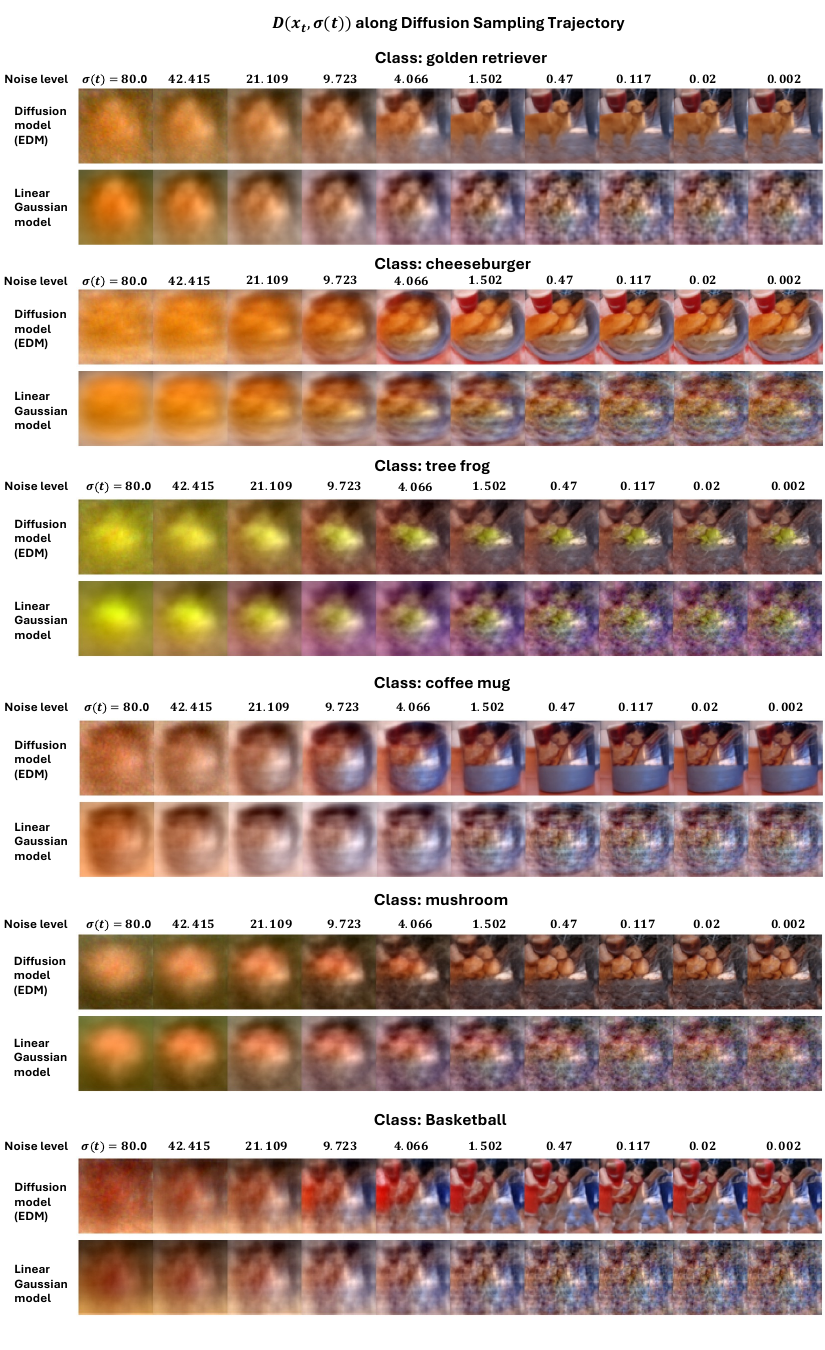}
\caption{\textbf{Similarity between Linear and Nonlinear Models.} For high to moderate noise levels ($\sigma(t)\in(4,80]$), the linear denoisers well approximate the learned deep denoisers. Though the two models diverge in lower noise reigmes, their final samples still match in overall structure. Although the linear models are built separately for each class according to~\eqref{linear Gaussian diffusion model}, they generate highly similar samples when starting from the same initial noise. The same similarity also exists in the samples of real-world diffusion models.}
\label{fig:similarity between lienar and nonlinear models}
\end{figure}

\subsection{Covariance Matrices of Different Classes Lack Class-Specificity} 
\label{subsec: Covariances matrices are not distinct enough}
The lack of class-specificity is especially pronounced in linear diffusion models. As shown in~\Cref{fig:various_steps_samples_visualization}(a) and~\Cref{fig:similarity between lienar and nonlinear models}, although the linear diffusion models are separately parameterized with the class-specific means and covariances for each class, the resulting samples share high similarity. Since the generated samples of the linear models are governed by the data covariances, the observed inter-class similarity implies that the covariance structures of different classes are not distinct enough.

\begin{figure}[t!]
    \centering
    \includegraphics[width=1\linewidth]{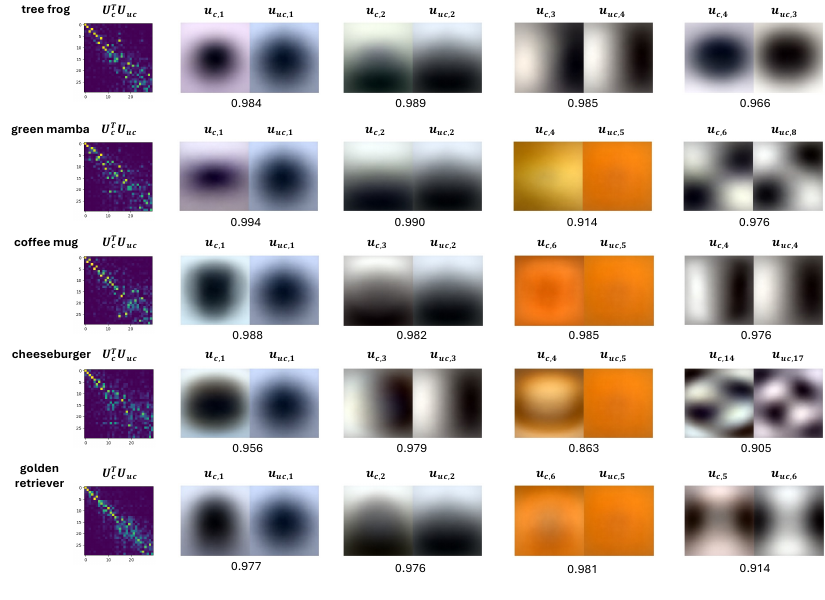}
    \caption{\textbf{Covariance Structures of CIFAR-10.} Each row corresponds to a different class. On the left, we show the correlation matrix between conditional and unconditional principal components (PCs), visualizing only the first 25. The subsequent images depict several highly correlated PCs, with correlation values displayed underneath. These results illustrate that the leading PCs do not always capture class-specific patterns.}
    \label{fig:covariance_structure_cifar}
\end{figure}

\begin{figure}[t!]
    \centering
    \includegraphics[width=1\linewidth]{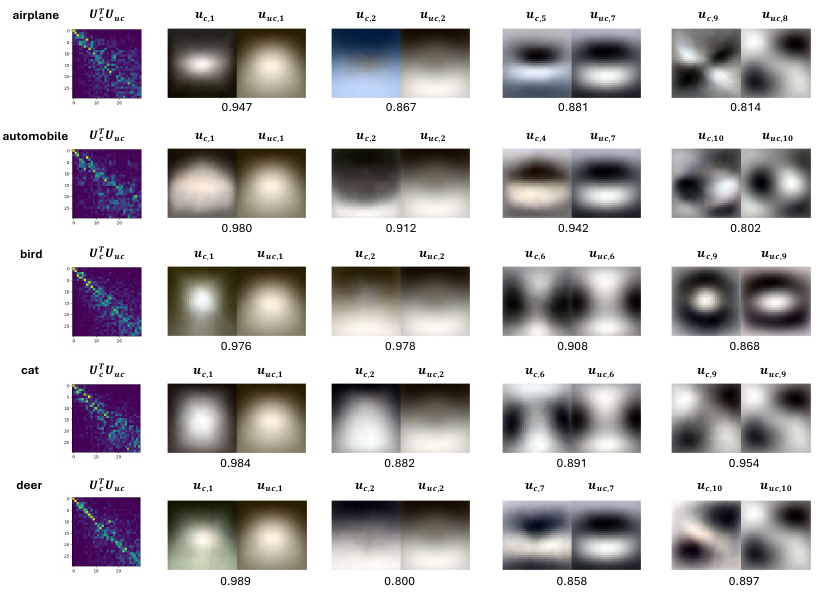}
    \caption{\textbf{Covariance Structures of ImageNet.} Each row corresponds to a different class. On the left, we show the correlation matrix between conditional and unconditional principal components (PCs), visualizing only the first 25. The subsequent images depict several highly correlated PCs, with correlation values displayed underneath. These results illustrate that the leading PCs do not always capture class-specific patterns.}
    \label{fig:covariance_structure_imagenet}
\end{figure}

Next, we quantitatively demonstrate that the class-specific covariance matrices are insufficiently distinct. To do this, we take $\mb U_{uc}$, the principal components (PCs) of the \emph{unconditional} dataset (i.e., the singular vectors of the unconditional covariance), as a baseline. We then compare $\mb U_{uc}$ to $\mb U_c$, the PCs of each \emph{conditional} dataset. As shown in~\Cref{fig:covariance_structure_cifar} and~\Cref{fig:covariance_structure_imagenet}, the correlation matrices $\mb U_c^T\mb U_{uc}$ for 10 classes (5 from CIFAR-10 and 5 from ImageNet) reveal that the leading PCs of each class share high similarity with those of the unconditional data. Thus, the PCs do not necessarily capture the distinctive features of individual classes though they represent the dominant variations of the dataset. Instead, these PCs often reflect global intensity or foreground-background variations.

\textbf{Why Covariance Structure Matters?} 
Covariance structures are fundamental statistics of a target distribution, and we would expect a robust diffusion model to learn them accurately. However, because these covariance structures are not sufficiently distinct, linear diffusion models—relying heavily on covariance for generation—struggle to produce high-quality images. To achieve better fidelity, models must leverage higher-order information beyond covariance. Recent works~\cite{liunderstanding,wangunreasonable} observe that deep diffusion models can be approximated \emph{unreasonably} well by linear diffusion models, especially when the model capacity is limited or the training is insufficient~\cite{liunderstanding}. Qualitatively, we have demonstrated the similarity between linear models and the actual diffusion models by showing that linear models replicate the coarse (low-frequency) features of samples generated by deep diffusion models. These results suggest that deep diffusion models may have an implicit bias toward learning simpler structures such as covariance, and thus the suboptimal nature of data covariance for generation task can limit their generative quality.


\section{Mechanism of Linear CFG}
\label{mechanism of linear CFG}
In the setting of linear diffusion model,~\eqref{term 2} and~\eqref{term 3} together form the CFG guidance. For the following discussion, we let $\Tilde{\mb \Sigma}_{c,t} = \mb U_c\Tilde{\mb\Lambda}_{\sigma(t),c}\mb U_c^T$ and $\Tilde{\mb \Sigma}_{uc,t} = \mb U_{uc}\Tilde{\mb\Lambda}_{\sigma(t),uc}\mb U_{uc}^T$.

\subsection{Mean-Shift Guidance}
\begin{figure}[t!]
    \centering
    \includegraphics[width=1\linewidth]{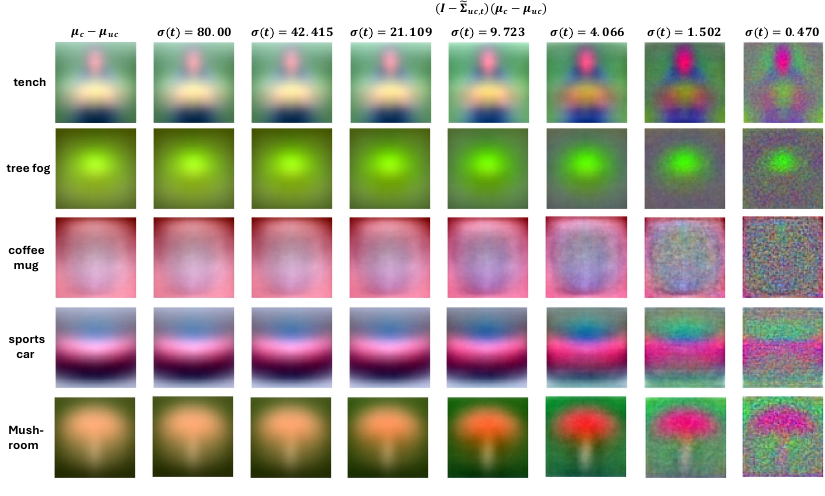}
    \caption{\textbf{Evolution of Mean-shift Guidance.} The leftmost image shows $\mb\mu_c-\mb\mu_{uc}$ while the subsequent images illustrate $(\mb I-\tilde{\mb \Sigma}_{uc,t})(\mb\mu_c-\mb\mu_{uc})$ at various noise levels $\sigma(t)$. Note that over a wide range of $\sigma(t)$, $(\mb I-\tilde{\mb \Sigma}_{uc,t})(\mb\mu_c-\mb\mu_{uc})$ remains close to $\mb\mu_c-\mb\mu_{uc}$.
    }
    \label{fig:evolution_of_mean_shift}
\end{figure}

\Cref{term 3} is the mean-shift guidance term that shifts $\mb x_t$ towards $(\mb I-\tilde{\mb \Sigma}_{uc,t})(\mb\mu_c-\mb\mu_{uc})$, a direction independent of the specific sample $\mb x_t$. At sufficiently large $\sigma(t)$, $(\mb I-\tilde{\mb\Sigma}_{uc,t})(\mb\mu_c-\mb\mu_{uc})\approx \mb\mu_c-\mb\mu_{uc}$, indicating the mean-shift term approximately shifts $\mb x_t$ towards the direction of the difference between class mean and unconditional mean $\mb\mu_c-\mb\mu_{uc}$. As $\sigma(t)$ decreases, the components of $\mb\mu_c-\mb\mu_{uc}$ within the subspace spanned by the unconditional PCs ($\mb U_{uc}$) progressively shrink to 0.~\Cref{fig:evolution_of_mean_shift} demonstrates $\mb \mu_c-\mb\mu_{uc}$ and the evolution of the mean-shift guidance term $(\mb I-\tilde{\mb \Sigma}_{uc,t})(\mb\mu_c-\mb\mu_{uc})$ across different noise levels. Notice that for a wide range of noise levels $\sigma(t)$, $(\mb I-\tilde{\mb \Sigma}_{uc,t})(\mb\mu_c-\mb\mu_{uc})$ remains close to $\mb \mu_c-\mb\mu_{uc}$, before it becomes uninformative at small $\sigma(t)$. Hence, as stated in the main text, the mean-shift guidance term has the effect of approximately shifting $\mb x_t$ in the direction $\mb \mu_c-\mb\mu_{uc}$.

\subsection{CPC guidance} 
\label{sec: CPC guidance appendix}
\begin{figure}[t!]
    \centering
    \includegraphics[width=1\linewidth]{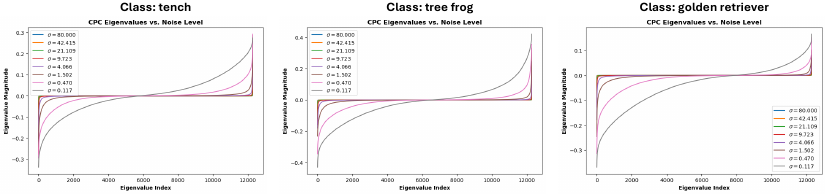}
    \caption{\textbf{Eigenvalues of $\Tilde{\mb \Sigma}_{c,t} - \Tilde{\mb \Sigma}_{uc,t}$.} The matrix $\Tilde{\mb \Sigma}_{c,t} - \Tilde{\mb \Sigma}_{uc,t}$ exhibits both positive and negative eigenvalues, whose corresponding eigenvectors correspond to positive and negative CPCs respectively. Though we only show the spectrum for three classes, this behavior remains consistent across other classes.}
    \label{fig:eigen spectrum}
\end{figure}

\begin{figure}[t!]
    \centering
    \includegraphics[width=1\linewidth]{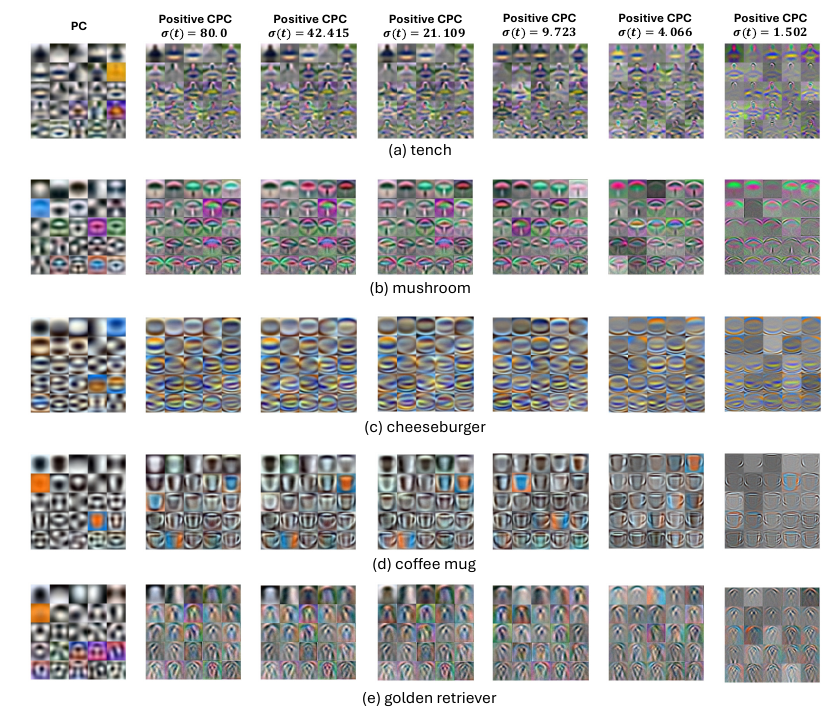}
    \caption{\textbf{Visualization of PCs and Positive CPCs.} Compared to principal components (PCs), the positive CPCs better capture class-specific patterns. Although only five classes are shown here, similar trends appear across other classes as well.}
    \label{fig: CPC_vs_PC_evolution}
\end{figure}

\Cref{term 2} is the CPC guidance term. Let $\mb V_{\sigma(t)}\hat{\mb\Lambda}_{\sigma(t)}\mb V_{\sigma(t)}^T$ be the eigendecomposition of 
$\Tilde{\mb \Sigma}_{c,t} - \Tilde{\mb \Sigma}_{uc,t}$, whose eigen spectrum is demonstrated in~\Cref{fig:eigen spectrum}, the CPC guidance term can be further decomposed into the positive CPC and negative CPC guidance:
\begin{align}
    \frac{\gamma}{\sigma^2(t)}(\mb V_{\sigma(t),+}\hat{\mb\Lambda}_{\sigma(t),+}\mb V^T_{\sigma(t),+})(\mb x_t-\mb\mu_c)\label{positive cpc term appendix}dt,\\\frac{\gamma}{\sigma^2(t)}(\mb V_{\sigma(t),-}\hat{\mb\Lambda}_{\sigma(t),-}\mb V^T_{\sigma(t),-})(\mb x_t-\mb\mu_c)dt\label{negative cpc term appendix},
\end{align}
where $\mb V_{\sigma(t),+}$ and $\mb V_{\sigma(t),-}$ contain eigenvectors corresponding to positive and negative eigenvalues $\hat{\mb\Lambda}_{\sigma(t),+}$ and $\hat{\mb\Lambda}_{\sigma(t),-}$ respectively. As discussed in \cref{posterior data covariance}, $\Tilde{\mb \Sigma}_{c,t}$ and $\Tilde{\mb \Sigma}_{uc,t}$ are (up to a factor $\sigma(t)^2$) the conditional and unconditional posterior covariances of $p_\text{data}(\mb x|\mb c)=\mathcal{N}(\mb \mu_c,\mb\Sigma_c)$ and $p_\text{data}(\mb x)=\mathcal{N}(\mb \mu_{uc}, \mb \Sigma_{uc})$. Hence, $\mb V_{\sigma(t)}$ are the CPCs which contrast between $X\sim p_{data}(\mb x|\mb x_t,\mb c)$ and $Y\sim p_\text{data}(\mb x|\mb x_t)$. Specifically, $\mb{V}_{\sigma(t),+}$ captures directions of higher conditional variance (class-specific features), while $\mb{V}_{\sigma(t),-}$ captures directions of higher unconditional variance (features more relevant to the unconditonal data).~\Cref{fig: CPC_vs_PC_evolution} illustrates the evolution of positive CPCs ($\mb{V}_{\sigma(t),+}$) and PCs ($\mb U_c$) across different noise levels. It is evident that the positive CPCs better capture the class-specific patterns compared to PCs. Here we choose not to display negative CPCs since they correspond to generic features that explain more variances for the unconditional dataset, which are less visually interpretable. Nevertheless, as we will show next, suppressing these directions is beneficial.

\subsection{Distinct Effects of the CFG Components} 
\label{Distinct Effects of the CFG Components appendix}
As we discussed in the main text, the three CFG components have the following effects respectively:
\begin{itemize}[leftmargin=*]
    \item The positive CPC guidance term amplifies the components of $\mb x_t$ that lie in the subspace spanned by the positive CPCs, thereby enhancing class-specific patterns.
    \item The negative CPC guidance term suppresses components of $\mb x_t$ that lie in the subspace spanned by the negative CPCs, mitigating background clutter and irrelevant content. As a result, the class-relevant sstructures become more salient. 
    \item The mean-shift term approximately shifts $\mb x_t$ in the direction $\mb\mu_c-\mb\mu_{uc}$, enhancing the structure of class mean within the generated samples. However, since this perturbation is independent of the specific $\mb x_t$, it tends to reduce sample diversity.
\end{itemize}

\begin{figure}[t!]
    \centering
    \includegraphics[width=0.9\linewidth]{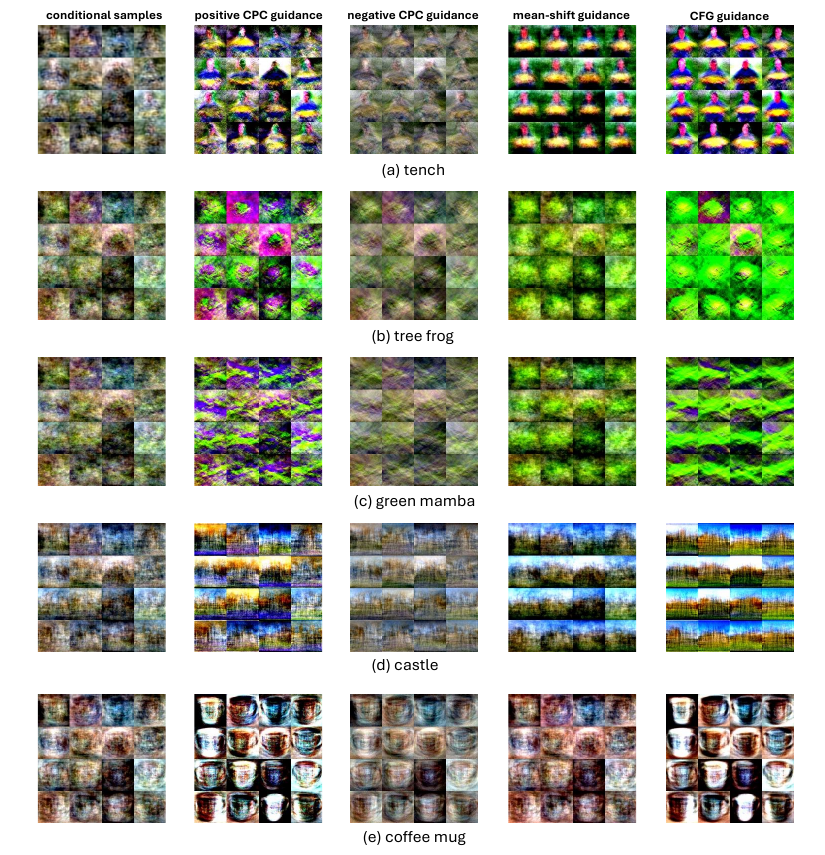}
    \caption{\textbf{Distinct Effects of Different CFG Components.} Each row shows (from left to right) the samples generated with (i) naive conditional sampling, (ii) guided with positive CPC term only (ii) guided with negative CPC term only, (iii) guided with mean-shift term only and (iv) guided with the full complete CFG. Each row corresponds to a different class.}
    \label{fig:linear CFG effects 1}
\end{figure}

\begin{figure}[t!]
    \centering
    \includegraphics[width=0.9\linewidth]{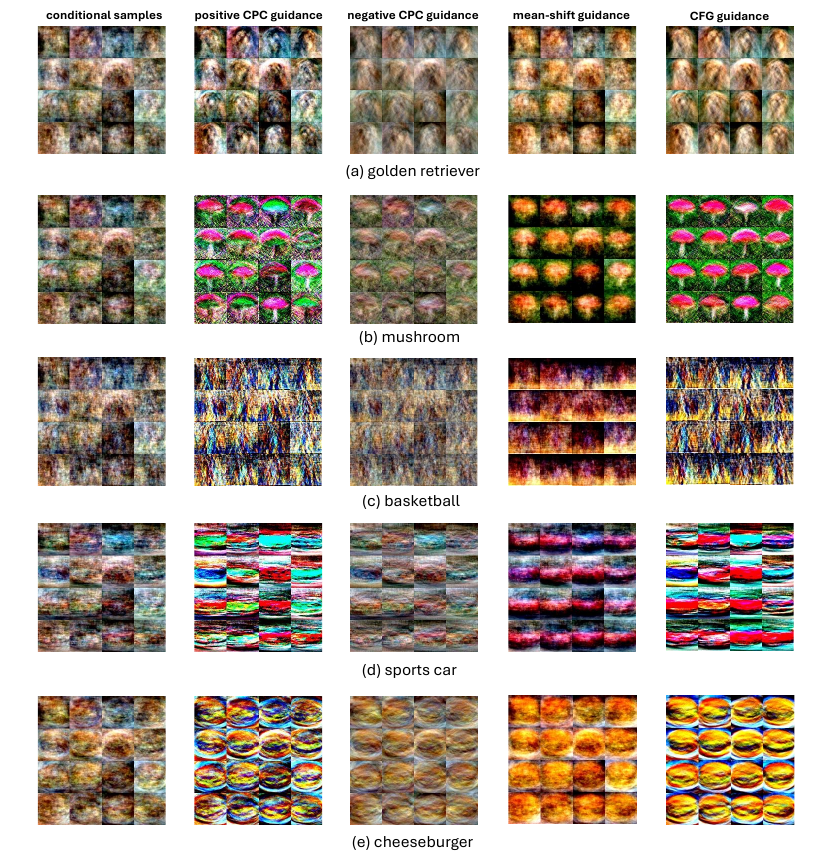}
    \caption{\textbf{Distinct Effects of Different CFG Components.} Each row shows (from left to right) the samples generated with (i) naive conditional sampling, (ii) guided with positive CPC term only (ii) guided with negative CPC term only, (iii) guided with mean-shift term only and (iv) guided with the full complete CFG. Each row corresponds to a different class.}
    \label{fig:linear cfg effects 2}
\end{figure}

\paragraph{Qualitative Results.}~\Cref{fig:linear CFG effects 1,fig:linear cfg effects 2} qualitatively demonstrates the effects of each CFG component in linear diffusion models over 10 different ImageNet classes.

\paragraph{Quantitative Results.}
The distortion effects of the CFG components can be quantitatively verified through the following experiment:
\begin{itemize}[leftmargin=*]
    \item For a chosen class, generate 1,000 samples using naive conditional sampling (denote the samples as $\mb x_c$) and 1,000 samples using CFG (denote the samples as $\mb x_\text{cfg}$), and center both sets by subtracting the class mean $\mb \mu_c$.
    \item For a chosen positive (or negative) CPC $\mb v$, compute the projection magnitudes $|\mb v^T(\mb x_c-\mb\mu_c)|$ and  $|\mb v^T(\mb x_\text{cfg}-\mb\mu_c)|$ to obtain a series of univariate distributions along $\mb v$.

    \item Project the same samples onto the mean-shift direction $\mb\mu_c-\mb\mu_{uc}$ by performing $(\mb\mu_c-\mb\mu_{uc})^T(\mb x_c-\mb\mu_c)$ and $(\mb\mu_c-\mb\mu_{uc})^T(\mb x_\text{cfg}-\mb\mu_c)$.
\end{itemize}

\begin{figure}[t]
    \centering
    \includegraphics[width=0.9\linewidth]{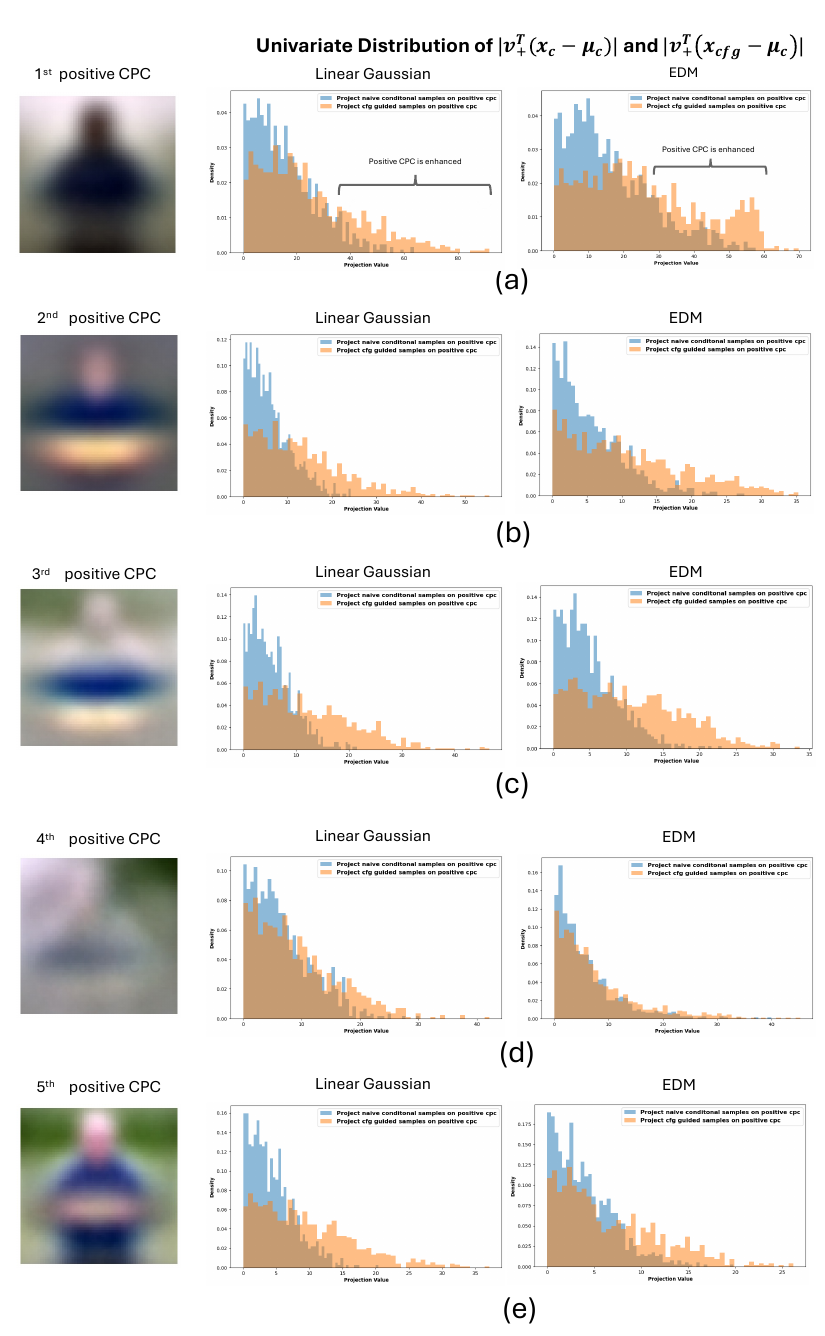}
    \caption{\textbf{CFG enhances positive CPCs.} For both linear and deep diffusion models, we randomly generate 1,000 naive conditional samples $\mb x_c$ and CFG-guided samples $\mb x_\text{cfg}$, center them by subtracting the class mean $\mb \mu_c$, and project them onto the top 5 positive CPCs ($\mb v_{+}$) to obtain a series univariate distributions. In both model types, the distributions of CFG-guided samples have greater density at higher projection values, suggesting that CFG amplifies the positive CPCs.}
    \label{fig:univariate distribution pos cpc}
\end{figure}

\begin{figure}[t!]
    \centering
    \includegraphics[width=0.9\linewidth]{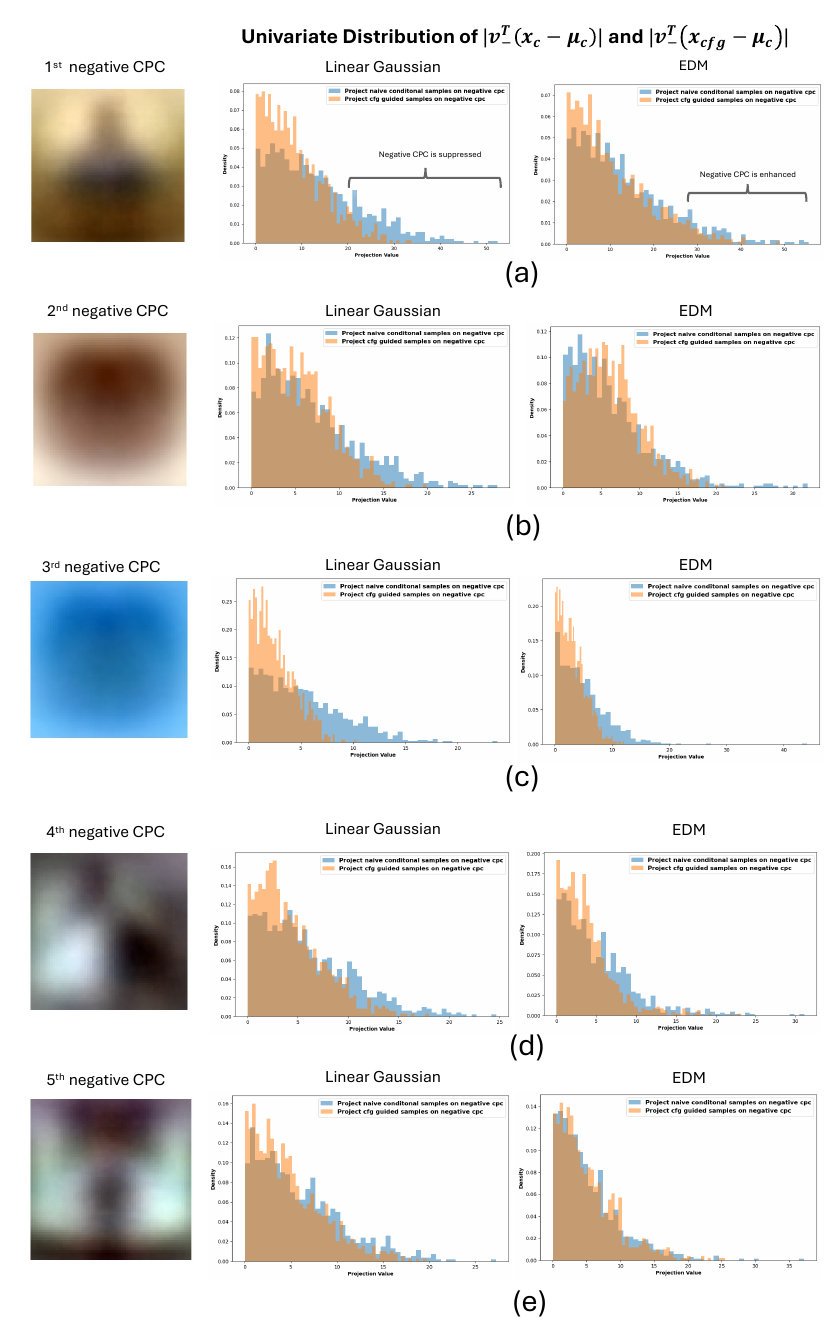}
    \caption{\textbf{CFG suppresses negative CPCs.} For both linear and deep diffusion models, we randomly generate 1,000 naive conditional samples $\mb x_c$ and CFG-guided samples $\mb x_\text{cfg}$, center them by subtracting the class mean $\mb \mu_c$, and project them onto the top 5 negative CPCs ($\mb v_{-}$) to obtain a series univariate distributions. In both model types, the distributions of CFG-guided samples have greater density at lower projection values, indicating that CFG suppresses the negative CPCs.}
    \label{fig:univariate distribution neg cpc}
\end{figure}

The resulting univariate distributions quantify the amount of energy the samples have along these directions. The above experiment are performed on both linear and nonlinear (EDM) diffusion models. The samples are generated using 20 steps and the guidance strength $\gamma$ is set to 2. We focus on the first class of ImageNet (tench) and present the results on the first 5 positive CPCs and negative CPCs. As shown in~\Cref{fig:univariate distribution pos cpc,fig:univariate distribution neg cpc}, compared to the samples with no CFG, the distributions of the CFG-guided samples have higher density on the positive CPC directions but lower density on the negative CPC directions, implying the former is enhanced while the latter is suppressed. The univariate distribution of the projection onto the mean-shift direction is presented in~\Cref{distinct effects of CFG on linear models}(c) (bottom row), from which it is clear that the density is shifted in the direction of $\mb\mu_c-\mb\mu_{uc}$.

\clearpage

\section{CFG in Nonlinear Deep Diffusion Models}
\label{CFG in Nonlinear deep diffusion model appendix}
In this section we provide additional experimental results for~\cref{investigate CFG in nonlinear models}, where we investigate how well the insights derived from linear diffusion models extend to real-world, nonlinear deep diffusion models. In this work, we study the state-of-the-art EDM models~\cite{karras2022elucidating}. 
\subsection{Linear to Nonlinear Transition in Diffusion Models}
\label{appendix linear to nonlinear transition}
\begin{figure}[t!]
    \centering
    \includegraphics[width=0.9\linewidth]{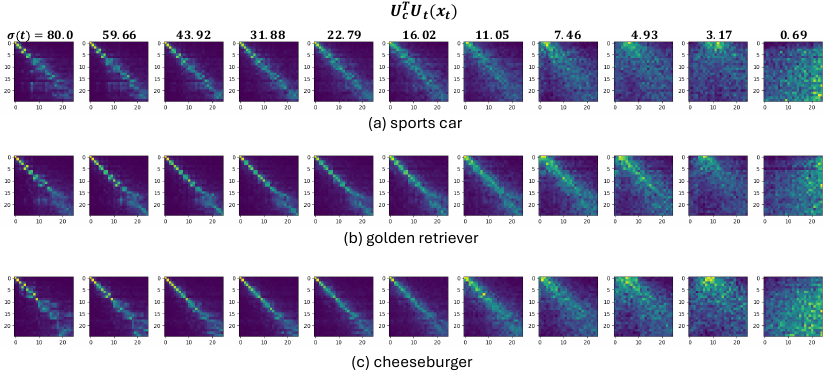}
    \caption{\textbf{Correlation between $\mb U_t(\mb x_t)$ and $\mb U_c$}.  The leading singular vectors of $\nabla\mathcal{D}_{\mb\theta}(\mb x_t;\sigma(t),\mb c)$ well align with $\mb U_c$ for high to moderate $\sigma(t)$. Each plot shows the average correlation computed 
    over 10 randomly initialized sampling trajectories measured for three different classes.} 
    \label{fig: correlation matrix three classes}
\end{figure}

\begin{figure}[t!]
    \centering
    \includegraphics[width=0.8\linewidth]{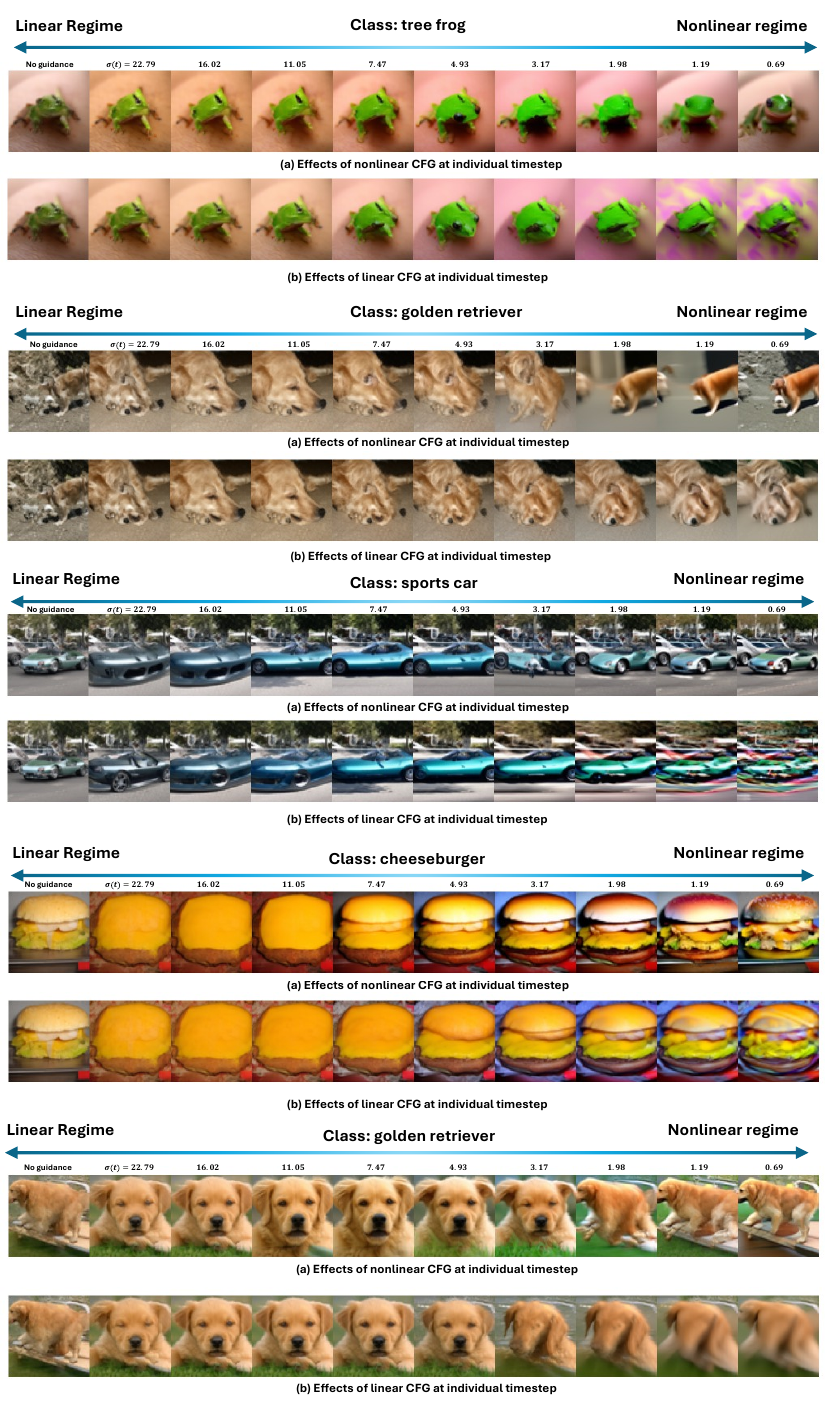}
    \caption{\textbf{Linear-to-nonlinear transition in diffusion models.}
    (a) and (b) compare nonlinear CFG and linear CFG applied to a deep diffusion model (EDM). The leftmost column shows unguided samples; subsequent columns show final samples when guidance is applied only at a specific noise level, with $\gamma=15$.}
    \label{fig:linear_nonlinear_transition_extra}
\end{figure}

\begin{figure}[t!]
    \centering
    \includegraphics[width=0.9\linewidth]{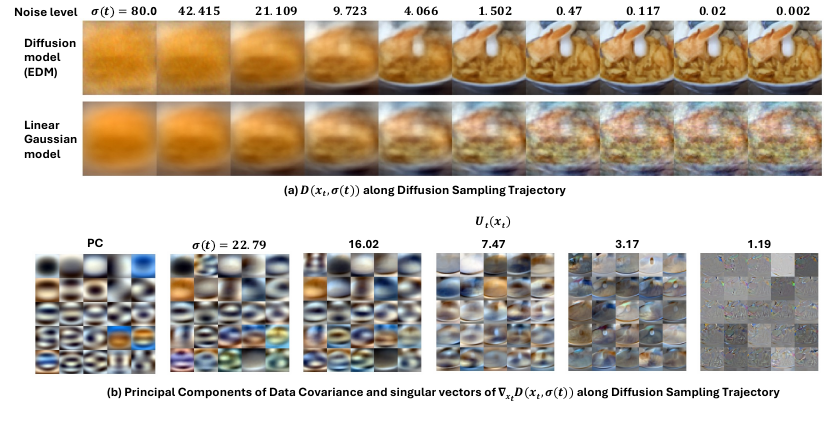}
    \caption{\textbf{Evolution of Denoiser Jacobian During Sampling}.(a) demonstrates one reverse diffusion trajectory. The left most image of (b) demonstrates the leading PCs of the data covariance. The subsequent images visualize the singular vectors, $\mb U_t(\mb x_t)$, of the denoiser Jacobian at different noise levels. Note that at early timesteps $\mb U_t(\mb x_t)$ match the PCs but gradually adapt to the geometry of the sample $\mb x_t$.} 
    \label{fig: Comparison between PC and Jacobian}
\end{figure}

Recent studies~\cite{liunderstanding,wangunreasonable} observe that at high to moderate noise levels, deep diffusion models $\mathcal{D}_{\mb\theta}(\mb x_t;\sigma(t))$ can be well approximated by the corresponding linear diffusion models $\mathcal{D}_{\mathrm{L}}(\mb x_t;\sigma(t))$ defined in~\eqref{linear Gaussian diffusion model}. As the noise level decreases, $\mathcal{D}_{\mb\theta}(\mb x_t;\sigma(t))$ becomes nonlinear. We verify this transition by the following experiment:

Let $\mb U_t(\mb x_t)$ be the left singular vectors of the network Jacobians $\nabla\mathcal{D}_{\mb\theta}(\mb x_t;\sigma(t),\mb c)$ along the sampling trajectories, and let $\mb U_c$ be the left singular vectors of $\nabla\mathcal{D}_{\mathrm{L}}(\mb x_t;\sigma(t),\mb c)$. Since $\mathcal{D}_{\mathrm{L}}(\mb x;\sigma (t)) = \mb \mu_c+\mb U_c\Tilde{\mb \Lambda}_{c,\sigma(t)}\mb U_c^T (\mb x-\mb \mu_c)$, if $\mathcal{D}_{\mb\theta}\approx \mathcal{D}_{\mathrm{L}}$, then $\nabla\mathcal{D}_{\mb\theta}(\mb x_t;\sigma(t),\mb c)\approx\nabla\mathcal{D}_{\mathrm{L}}(\mb x_t;\sigma(t),\mb c)\approx \mb U_c\Tilde{\mb \Lambda}_{c,\sigma(t)}\mb U_c^T$, implying $\mb U_t(\mb x_t)\approx \mb U_c$, independent of $\mb x_t$. As illustrated in~\Cref{fig: correlation matrix three classes}, for large $\sigma(t)$, the leading singular vectors of $\nabla \mathcal{D}_{\mb\theta}(\mb x_t;\sigma(t),\mb c)$ indeed align with $\mb U_c$. Note that since $\Tilde{\mb \Lambda}_{c,\sigma(t)}=\text{diag}(\frac{\lambda_{c,1}}{\lambda_{c,1}+\sigma^2(t)},...,\frac{\lambda_{c,d}}{\lambda_{c,d}+\sigma^2(t)})$, $\nabla\mathcal{D}_{\mathrm{L}}(\mb x;\sigma (t))$ is highly low-rank at large $\sigma(t)$. Thus, our primary interest is in the leading singular vectors, and the non-leading singular vectors are ambiguous. In contrast, for small $\sigma(t)$, the alignment no longer holds and $\nabla \mathcal{D}_{\mb\theta}(\mb x_t;\sigma(t))$ starts to adapt to individual samples, reflecting the model's nonlinear behavior.~\Cref{fig:linear_nonlinear_transition,fig:linear_nonlinear_transition_extra,fig: Comparison between PC and Jacobian} qualitatively demonstrates this linear to nonlinear transition.


\subsection{CFG in the Linear Regime}
\label{CFG linear reimge discussion appendix}
\begin{figure}
    \centering
    \includegraphics[width=1\linewidth]{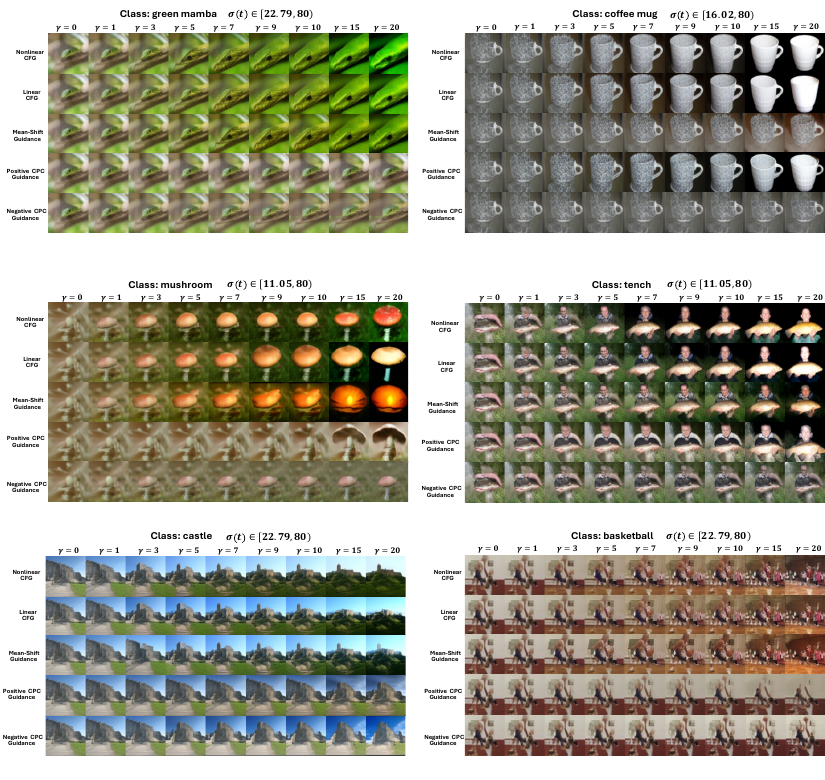}
    \caption{\textbf{Effects of CFG in Linear Regime.} Each row demonstrates the impact of different guidance types applied to EDM within the linear regime, with varying guidance strength $\gamma$. The guidance is applied only within intervals specified in the subtitles, where the model exhibits linear behavior.}
    \label{fig:linear_regime_qualitative_extra}
\end{figure}

\begin{figure}
    \centering
    \includegraphics[width=1\linewidth]{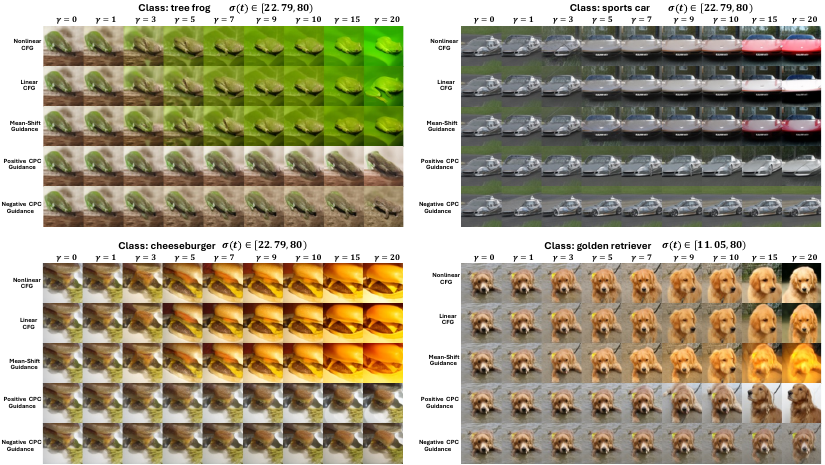}
    \caption{\textbf{Effects of CFG in Linear Regime.} Each row demonstrates the impact of different guidance types applied to EDM within the linear regime, with varying guidance strength $\gamma$. The guidance is applied only within intervals specified in the subtitles, where the model exhibits linear behavior.}
    \label{fig:linear_regime_qualitative_extra_2}
\end{figure}

\begin{figure}
    \centering
    \includegraphics[width=1\linewidth]{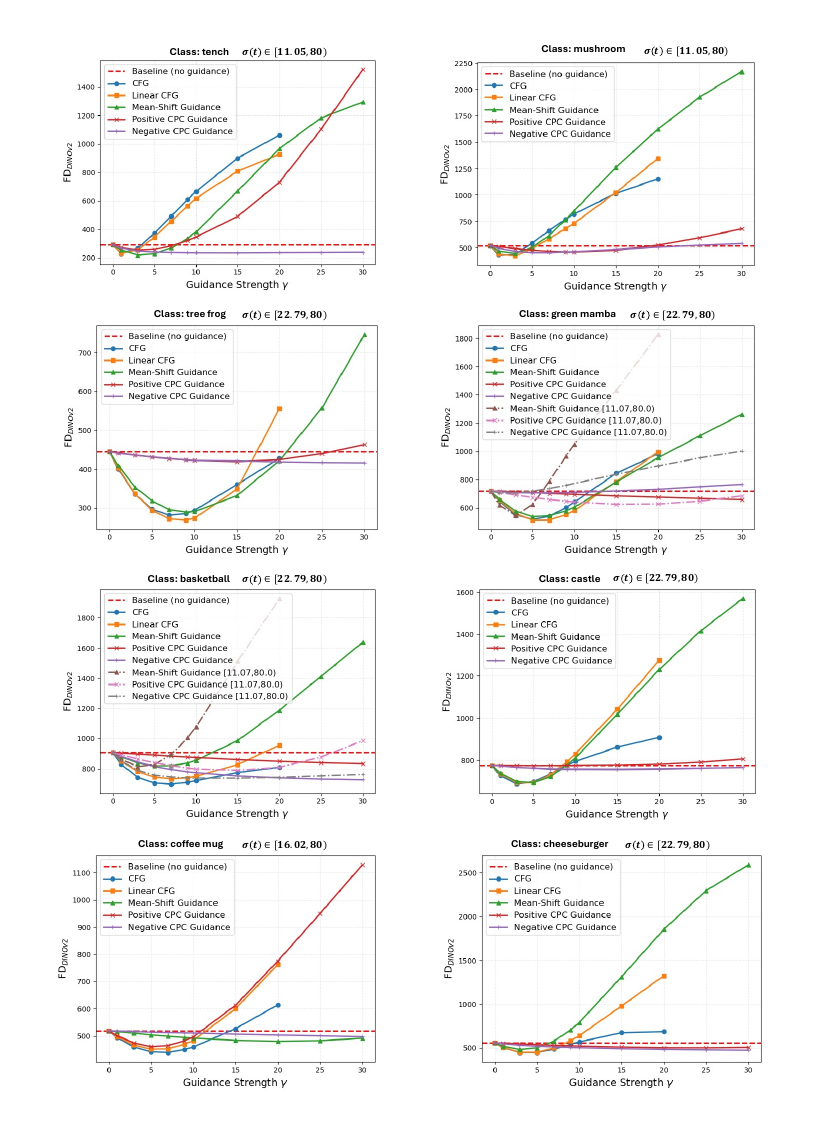}
    \caption{\textbf{FD\textsubscript{DINOv2} Scores.} The guidance is applied to the interval specified in the subtitles. For green mamba and basketball, we find it beneficial to apply CPC guidance beyond the linear regime, with results demonstrated by the dashed curves.}
    \label{fig:linear_regime_quantatitive_extra}
\end{figure}

\begin{figure}[h]
    \centering
    \includegraphics[width=1\linewidth]{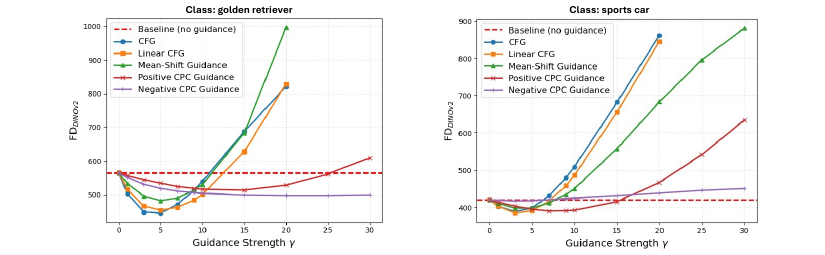}
    \caption{\textbf{FD\textsubscript{DINOv2} Scores.} The reported values are relatively high because the scores are computed separately per class, which often has a limited number of training images. It is well known that FD\textsubscript{DINOv2} scores can appear inflated when the reference dataset size is small. }
    \label{fig:FD_score}
    \vspace{-0.1in}
\end{figure}

We provide additional experimental results for~\cref{linear regime} in~\Cref{fig:linear_regime_qualitative_extra,fig:linear_regime_quantatitive_extra,fig:linear_regime_qualitative_extra_2,fig:FD_score}. Because the precise transition time from the linear to the nonlinear regime—as well as the influence of each CFG component—varies across classes, we empirically choose the interval for applying guidance and calculate the FD\textsubscript{DINOv2} score with 50,000 generated images for each class separately. We summarize our observations as follows (see also~\cref{linear regime}):
\begin{itemize}[leftmargin=*]
\item \textbf{Linear vs.\ Nonlinear CFG.} Applying linear CFG to deep diffusion models produces effects that closely resemble those of the actual (nonlinear) CFG. \item \textbf{Dominance of Mean-Shift.} In most of the 10 classes studied, the mean-shift guidance term dominates CFG behavior, as it alone can generate results visually similar to full CFG. However, for the coffee mug class, the positive CPC term takes precedence, becoming the primary driver of CFG. \item \textbf{Role of CPC Guidance.} CPC guidance generally improves generation quality, though its benefits can sometimes be less pronounced. For instance, in the tree frog and castle classes (\Cref{fig:linear_regime_quantatitive_extra}), the CPC term does not enhance FD\textsubscript{DINOv2} as much as the mean-shift term. Nevertheless, CPC guidance operates effectively over a wider range of guidance strengths $\gamma$ and noise intervals. For the green mamba and basketball classes, we show results within the prescribed noise interval as solid curves, and extend beyond this interval as dashed curves. While mean-shift becomes highly detrimental once outside the linear regime, CPC guidance remains beneficial. 
\end{itemize}

\subsection{Mean-Shifted Noise Initialization}
\label{Mean-shifted noise initialization trick}
The observation that the sample-independent mean-shift guidance alone leads to improved FD\textsubscript{DINOv2} score implies that simply initializing the sampling process from a mean-shifted Gaussian, $\mb x_T\sim \mathcal{N}(\gamma(\mb\mu_c-\mb\mu_{uc}),\sigma^2(T)\mb I)$, with no additional guidance applied, can improve the generation quality, which we verify through the following experiment:
\begin{itemize}[leftmargin=*]
    \item For a chosen class and a positive scalar $\gamma$, generate 50,000 samples via naive conditional sampling initialized from a mean-shifted Gaussian $\mathcal{N}(\gamma(\mb\mu_c-\mb\mu_{uc}),\sigma^2(T)\mb I)$. Then evaluate the sample quality with FID and FD\textsubscript{DINOv2} scores.
    \item Repeat the above procedure across several classes and a range of $\gamma$ values.
\end{itemize}

\begin{figure}
    \centering
    \includegraphics[width=0.8\linewidth]{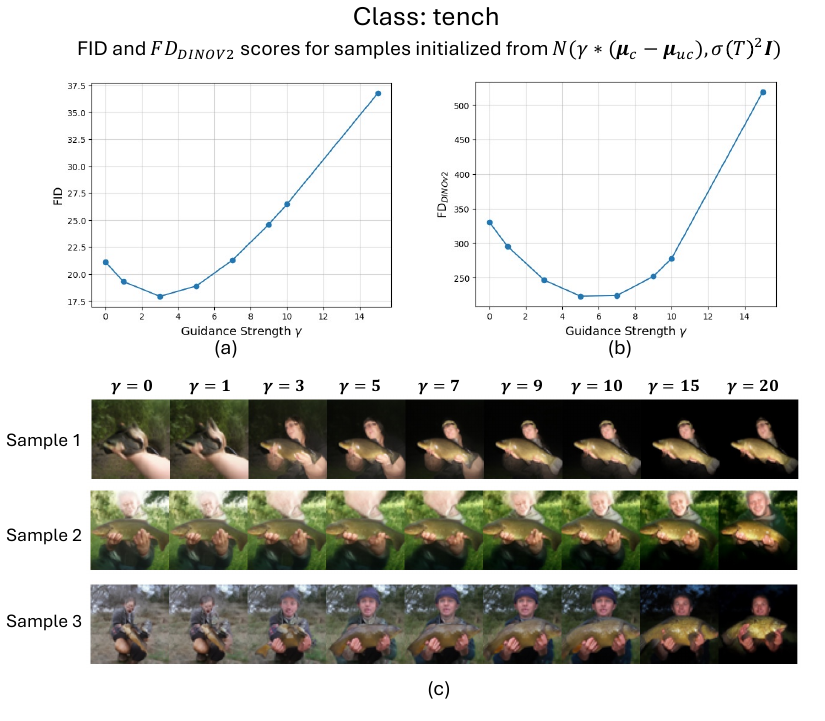}
    \caption{\textbf{Effects of initializing with mean-shift.} For every $\gamma\in[0,1,3,5,7,9,10,15,20]$, we generate 50,000 images from initial nosies sampled from mean-shifted Gaussian distribution $\mathcal{N}(\gamma(\mb\mu_c-\mb\mu_{uc}),\sigma(T)^2\mb I)$ and compute the FID scores (a) and $\text{FD}_{DINOV2}$ scores (b). The samples are visualized in (c). Note that adding mean-shift to the initial distribution leads to improvement of standard metrics.}
    \label{initialization trick 1}
\end{figure}

\begin{figure}
    \centering
    \includegraphics[width=0.8\linewidth]{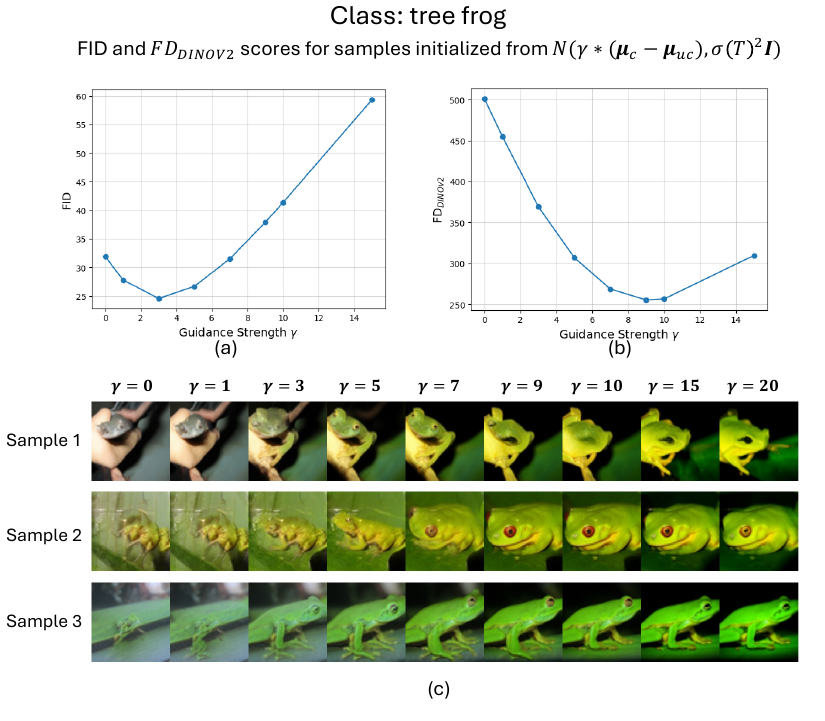}
    \caption{\textbf{Effects of initializing with mean-shift.} For every $\gamma\in[0,1,3,5,7,9,10,15,20]$, we generate 50,000 images from initial nosies sampled from mean-shifted Gaussian distribution $\mathcal{N}(\gamma(\mb\mu_c-\mb\mu_{uc}),\sigma(T)^2\mb I)$ and compute the FID scores (a) and $\text{FD}_{DINOV2}$ scores (b). The samples are visualized in (c). Note that adding mean-shift to the initial distribution leads to improvement of standard metrics.}
    \label{initialization trick 2}
\end{figure}

\begin{figure}
    \centering
    \includegraphics[width=0.8\linewidth]{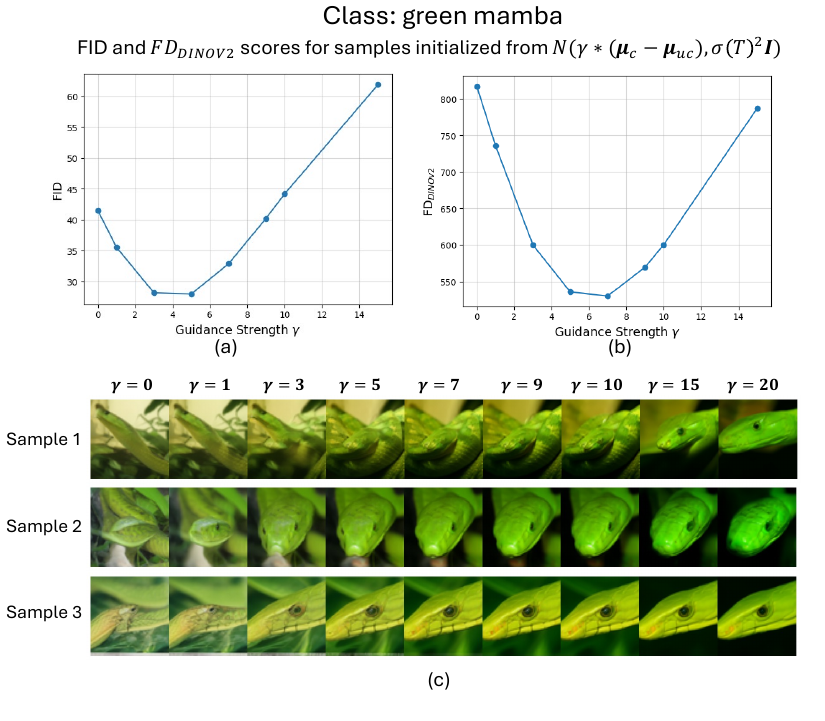}
    \caption{\textbf{Effects of initializing with mean-shift.} For every $\gamma\in[0,1,3,5,7,9,10,15,20]$, we generate 50,000 images from initial nosies sampled from mean-shifted Gaussian distribution $\mathcal{N}(\gamma(\mb\mu_c-\mb\mu_{uc}),\sigma(T)^2\mb I)$ and compute the FID scores (a) and $\text{FD}_{DINOV2}$ scores (b). The samples are visualized in (c). Note that adding mean-shift to the initial distribution leads to improvement of standard metrics.}
    \label{initialization trick 3}
\end{figure}

\begin{figure}
    \centering
    \includegraphics[width=0.8\linewidth]{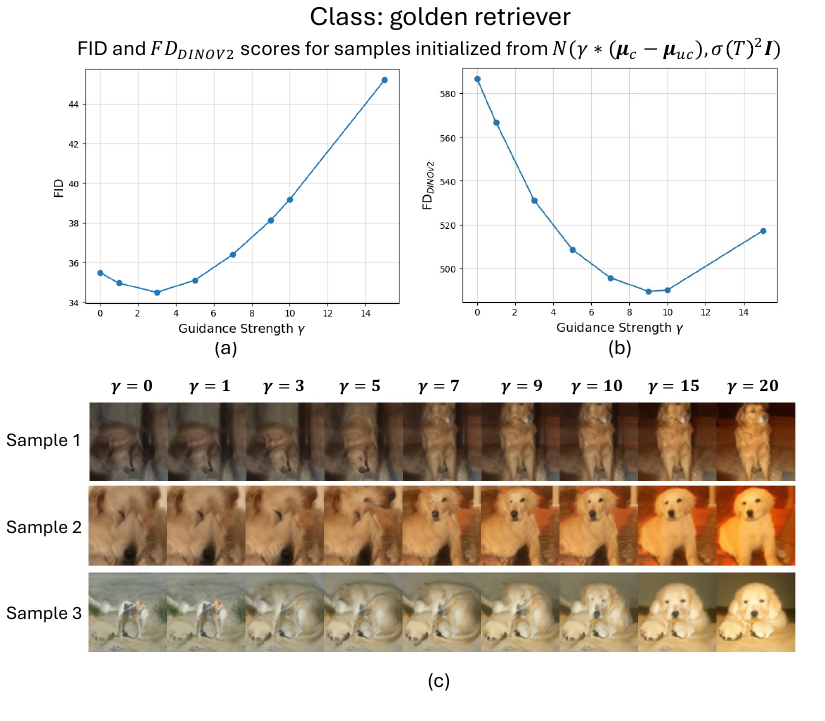}
    \caption{\textbf{Effects of initializing with mean-shift.} For every $\gamma\in[0,1,3,5,7,9,10,15,20]$, we generate 50,000 images from initial nosies sampled from mean-shifted Gaussian distribution $\mathcal{N}(\gamma(\mb\mu_c-\mb\mu_{uc}),\sigma(T)^2\mb I)$ and compute the FID scores (a) and $\text{FD}_{DINOV2}$ scores (b). The samples are visualized in (c). Note that adding mean-shift to the initial distribution leads to improvement of standard metrics.}
    \label{initialization trick 4}
\end{figure}

\begin{figure}
    \centering
    \includegraphics[width=0.8\linewidth]{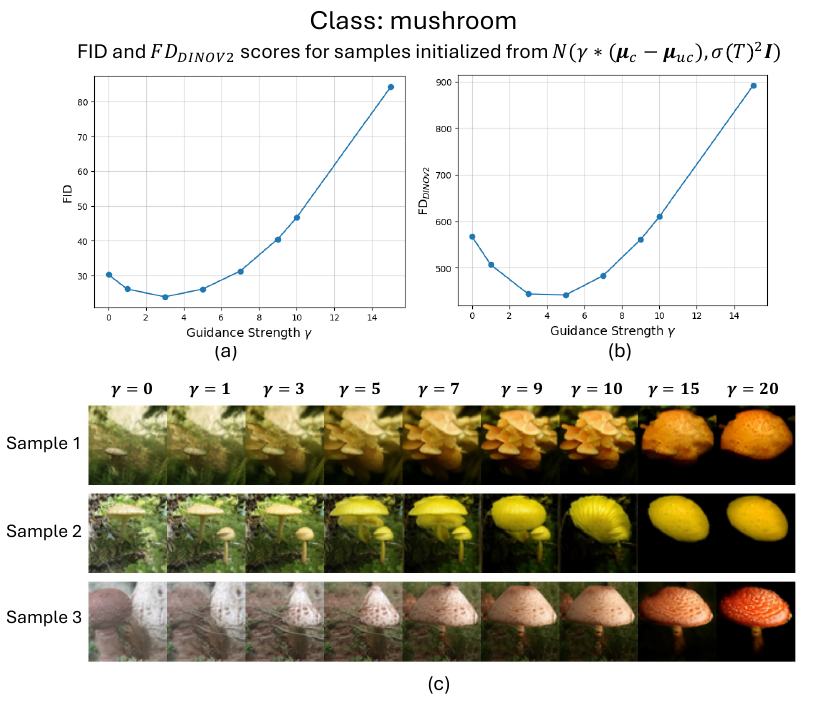}
    \caption{\textbf{Effects of initializing with mean-shift.} For every $\gamma\in[0,1,3,5,7,9,10,15,20]$, we generate 50,000 images from initial nosies sampled from mean-shifted Gaussian distribution $\mathcal{N}(\gamma(\mb\mu_c-\mb\mu_{uc}),\sigma(T)^2\mb I)$ and compute the FID scores (a) and $\text{FD}_{DINOV2}$ scores (b). The samples are visualized in (c). Note that adding mean-shift to the initial distribution leads to improvement of standard metrics.}
    \label{initialization trick 5}
\end{figure}

We perform the above experiments on 5 classes, where $\sigma(T)$ is set to 31.9. The results are shown in~\Cref{initialization trick 1,initialization trick 2,initialization trick 3,initialization trick 4,initialization trick 5}. Note that the sample quality improves with a properly chosen $\gamma$.

\subsection{CFG in the Nonlinear Regime}
\label{CFG in Nonlinear Regime appendix}
\begin{figure}[t!]
    \centering
    \includegraphics[width=1\linewidth]{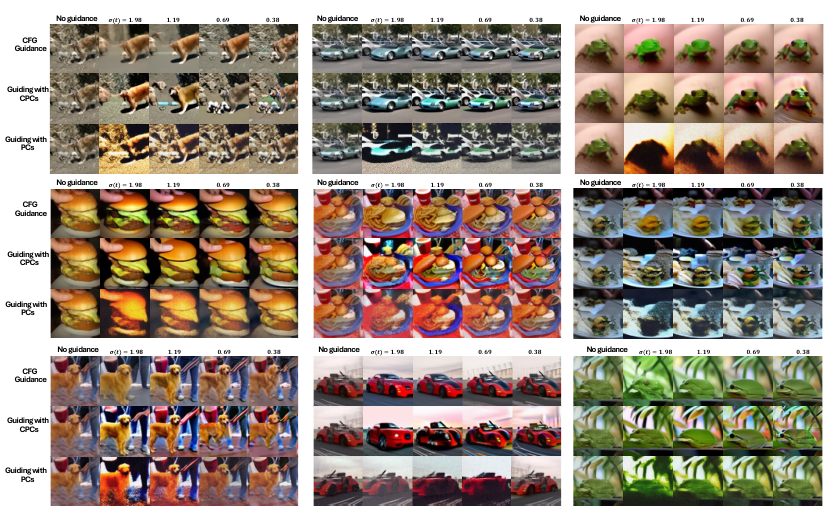}
    \caption{\textbf{Effects of CFG in the Nonlinear Regime.} Different guidance methods, each with a fixed strength of $\gamma=15$, are applied at individual timesteps in the nonlinear regime. Each image shows the final output when guidance is applied solely at the timestep indicated at the top. Note that \eqref{scaled positive cpc subspace projection Jacobian} closely matches the effects of CFG by enhancing finer image details, whereas \eqref{scaled positive PCA projection Jacobian} does not improve generation quality.} 
    \label{fig:construct_nonlinear_CFG_extra}
\end{figure}

We provide additional experimental results for \cref{sec: nonlienar regime investigation} in \Cref{fig:construct_nonlinear_CFG_extra}. We argue that effective guidance in this regime should satisfy two key properties:

\begin{itemize}[leftmargin=*]
    \item \textbf{Capture local structure of a specific sample.} As shown in \Cref{fig:linear_nonlinear_transition_extra}, when $\sigma(t)$ is small, the model diverges considerably from its linear approximation, and linear CFG deviates from the actual nonlinear CFG. In this regime, CFG does not alter the overall image structure but instead refines existing features to produce crisper images. Consequently, effective guidance must adapt to each specific sample. We propose that such guidance can be derived from the network Jacobians $\nabla\mathcal{D}_{\mb\theta}(\mb x_t;\sigma(t),\mb c)$ evaluated at $\mb x_t$. Prior work~\cite{kadkhodaiegeneralization} shows that the singular vectors of these Jacobians, which are equivalent to the posterior covariances, adapt to the input $\mb x_t$.
    
    \item \textbf{Capture class-specific patterns.} As in the linear case, the guidance must also capture class-specific patterns. This can be achieved by contrasting the conditional Jacobian $\nabla\mathcal{D}_{\mb\theta}(\mb x_t;\sigma(t),\mb c)$ with the unconditional Jacobian $\nabla\mathcal{D}_{\mb\theta}(\mb x_t;\sigma(t))$. \Cref{fig:construct_nonlinear_CFG_extra} shows that guidance built using CPCs—i.e., the difference between these two Jacobians—yields effects similar to actual CFG. In contrast, guidance derived solely from the conditional Jacobian does not improve image quality.
\end{itemize}

\noindent
Note that \eqref{scaled positive cpc subspace projection Jacobian} is inspired by linear positive CPC guidance~\eqref{scaled positive cpc subspace projection}. We also test other guidance such as
\begin{align}
    \frac{\gamma}{\sigma^2(t)}\sum_i \hat{\lambda}_{+,i}\,\mb v_{+,i}\,(\mb v_{+,i}^T(\mb x_t-\mb\mu_c)),
\end{align}
but find it less effective than \eqref{scaled positive cpc subspace projection Jacobian}, likely due to additional noise in $\mb x_t$. Moreover, we observe that negative CPCs and mean-shift terms are not as effective in the nonlinear regime.

Lastly, we'd like to remark that our goal here is \emph{not} to suggest that CFG in the nonlinear regime is exactly equivalent to \eqref{scaled positive cpc subspace projection Jacobian}; rather, we note that both approaches exhibit similar behaviors, implying they may share a core mechanism: identifying and amplifying \emph{sample-specific} and \emph{class-specific} features. The exact analytical form of CFG in the nonlinear setting remains challenging to derive due to the complexity of deep networks, leaving a promising direction for future work.

\clearpage
\section{Experimental Results on Latent Diffusion Models}
\label{EDM-2 experiments}
In the main text, we conducted experiments using the EDM-1 model~\cite{karras2022elucidating}, which operates directly in pixel space with $64\times64$ resolution. Here, we present complementary results on the EDM-2~\cite{karras2024analyzing} latent diffusion model, which generates images at $512\times512$ resolution.

\begin{figure}[t!]
    \centering
    \includegraphics[width=1\linewidth]{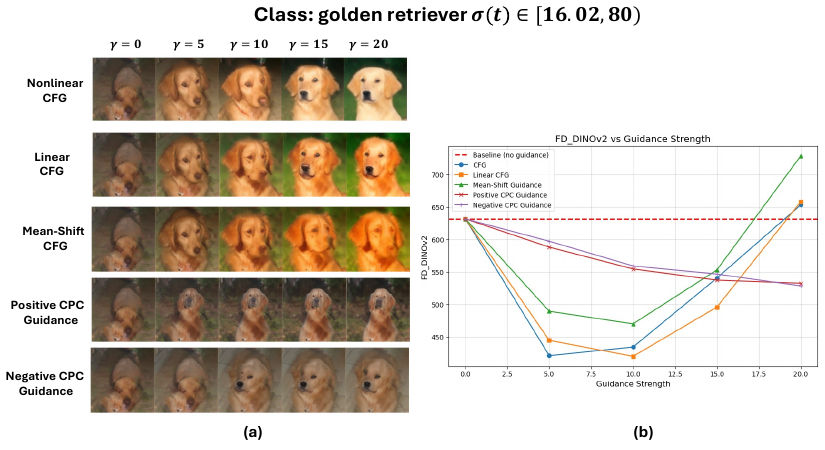}
    \caption{\textbf{Effects of CFG in the Linear Regime (EDM-2).} Each row in (a) demonstrates the impact of different guidance types applied to EDM-2 within the linear regime (specified in the subtitles), with varying guidance strength $\gamma$. (b) shows the FD\textsubscript{DINOv2} scores computed over 50,000 samples.} 
    \label{fig:EDM_2_linear_regime}
\end{figure}

\textbf{Linear Regime.} We evaluate multiple guidance strategies—including actual CFG, linear CFG, Mean-shift guidance, positive CPC guidance, and negative CPC guidance—within the high-noise intervals (the linear regime). For each method, we generate 50,000 images conditioned on the class label \textit{“golden retriever”} and compute the $\mathrm{FD_{DINOv2}}$ metric. The results, shown in~\Cref{fig:EDM_2_linear_regime}, are consistent with the observations reported in the main text.

\begin{figure}[t!]
    \centering
    \includegraphics[width=0.8\linewidth]{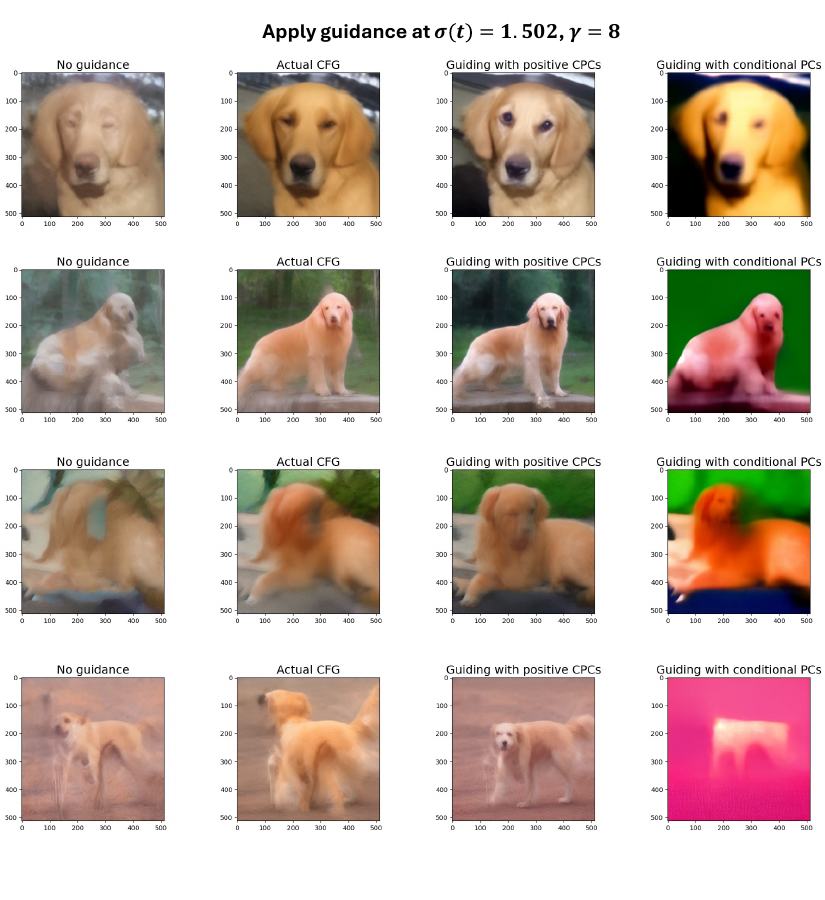}
    \vspace{-0.1in}
    \caption{\textbf{Effects of CFG in Nonlinear Regime (EDM-2).} Different guidance methods, each with a fixed strength of $\gamma=8$, are applied at $\sigma(t)=1.502$. The samples in each row are generated from the same initial noise.} 
    \label{fig:EDM_2_nonlinear_regime}
\end{figure}

\textbf{Nonlinear Regime.} We next examine guidance effects in the nonlinear regime using~\eqref{scaled positive cpc subspace projection Jacobian} and~\eqref{scaled positive PCA projection Jacobian}. As shown in~\Cref{fig:EDM_2_nonlinear_regime}, guiding with CPCs produces visual effects similar to those of actual CFG—enhancing image sharpness and structure—whereas guidance with conditional PCs often leads to oversaturated colors. This highlights the importance of selectively amplifying class-specific features. We note that our heuristic guidance serves as a conceptual approximation and may not always perfectly align with actual CFG behavior. Additional failure cases will be provided in our code release.

\clearpage
\section{CFG in Gaussian Mixture Model}
\label{analyze CFG in Gaussian Mixtures}
Thus far we've been focusing on the setting of linear diffusion models, in which the learned score functions are equivalent to those of a Multivariate Gaussian distribution. From a complementary perspective, several works~\cite{chidambaramdoes,wutheoretical,bradley2024classifier} have studied CFG under the Gaussian mixture model data assumption. However, these works assume each Gaussian cluster has isotropic covariance, which is oversimplified for natural image dataset. In this section, we demonstrate that CFG guidance under Gaussian mixture model can be decomposed in a similar way as the case of linear diffusion model. 

Consider unconditional data distribution:
\begin{align}
    p_\text{data}(\mb x)=\sum_{i=1}^K\pi_i\mathcal{N}(\mb x;\mb \mu_i,\mb\Sigma_i),
\end{align}
where $\mb\mu_i$ and $\mb\Sigma_i$ are the mean and covariance of the $i^{th}$ cluster with weight $\pi_i$. The noise-mollified data distribution then takes the following form:
\begin{align}
    p(\mb x;\sigma(t))=\sum_{i=1}^K\pi_i\mathcal{N}(\mb x;\mb \mu_i,\mb\Sigma_i+\sigma^2(t)\mb I).
\end{align}
Let $\mb\Sigma_{\sigma(t),i}:=\mb\Sigma_i+\sigma^2(t)\mb I$, then the score function of $p(\mb x;\sigma(t))$ is:
\begin{align}
    \nabla \log p(\mb x;\sigma(t))&=\frac{\nabla p(\mb x;\sigma(t))}{p(\mb x;\sigma(t))}\\
                                  &= \frac{\sum_{i=1}^K\pi_i\nabla\mathcal{N}(\mb x;\mb\mu_i,\mb\Sigma_{\sigma(t),i})}{\sum_{i=1}^K\pi_i\mathcal{N}(\mb x;\mb\mu_i,\mb\Sigma_{\sigma(t),i})}\\
                                  &=\frac{\sum_{i=1}^K\pi_i\mathcal{N}(\mb x;\mb\mu_i,\mb\Sigma_{\sigma(t),i})\mb\Sigma_{\sigma(t),i}^{-1}(\mb\mu_i-\mb x)}{\sum_{i=1}^K\pi_i\mathcal{N}(\mb x;\mb\mu_i,\mb\Sigma_{\sigma(t),i})}\\
                                  &=\sum_{i=1}^K w_i(\mb x)\mb \Sigma_{\sigma(t),i}^{-1}(\mb\mu_i-\mb x),
\end{align}
where $w_i(\mb x)=\frac{\pi_i\mathcal{N}(\mb x;\mb\mu_i,\mb\Sigma_{\sigma(t),i})} {\sum_{i=1}^K\pi_i\mathcal{N}(\mb x;\mb\mu_i,\mb\Sigma_{\sigma(t),i})}$ representing the posterior probability that $\mb x$ belongs to the $i^{th}$ cluster and $\sum_{i=1}^Kw_i(\mb x)=1$. Let $\mb\Sigma_i=\mb U_i\mb\Lambda_i\mb U_i^T$ be the full SVD where $\mb\Lambda_i=\text{diag}(\lambda_{i,1},\cdots,\lambda_{i,d})$, by Tweedie's formula, the optimal denoiser of the noise-mollified Gaussian mixture model takes the following form:
\begin{align}
    \mathcal{D}(\mb x;\sigma(t))&=\mb x+\sigma^2(t)\nabla\log p(\mb x;\sigma(t))\\
    &=\mb x+\sigma^2(t)\sum_{i=1}^Kw_i(\mb x)\mb \Sigma_{\sigma(t),i}^{-1}(\mb\mu_i-\mb x)\\
    &=\sum_{i=1}^K w_i(\mb x)\mb\mu_i+\sum_{i=1}^Kw_i(\mb x)\mb U_i\mb\tilde{\mb\Lambda}_{\sigma(t),i}\mb U_i^T(\mb x-\mb\mu_i),
\end{align}
where $\tilde{\mb\Lambda}_{\sigma(t),i}=\diag \paren{\frac{\lambda_{i,1}}{\lambda_{i,1}+\sigma^2(t)}, \cdots, \frac{\lambda_{i,d}}{\lambda_{i,d}+\sigma^2(t)} }$. Furthermore, under the Gaussian mixture model assumption, each conditional distribution is a Gaussian distribution and from~\eqref{linear Gaussian diffusion model} we know the conditional optimal denoiser of the $i^{th}$ cluster is:
\begin{align}
    \mathcal{D}(\mb x;\sigma(t),\mb c_i)=\mb \mu_i+\mb U_i\Tilde{\mb \Lambda}_{\sigma(t),i}\mb U_i^T (\mb x-\mb \mu_i).
\end{align}
Without loss of generality, we set the target condition as $\mb c_1$. Then the CFG guidance at timestep $t$ takes the form:
\begin{align}
    g(\mb x, t)&=\nabla_{\mb x_t}\log p(\mb x|\mb c_1;\sigma(t))-\nabla_{\mb x_t}\log p(\mb x;\sigma(t))\\
        &=\frac{1}{\sigma^2(t)}(\mathcal{D}(\mb x_t;\sigma(t),\mb c_1)-\mathcal{D}(\mb x_t;\sigma(t)))\\
        \label{mixture cpc}
        &=\frac{1}{\sigma^2(t)}(\mb U_1\tilde{\mb\Lambda}_{\sigma(t),1}\mb U_1^T-\sum_{i=1}^Kw_i(\mb x)\mb U_i\tilde{\mb\Lambda}_{\sigma(t),i}\mb U_i^T)(\mb x-\mb\mu_1)\\
        \label{mixture mean-shift}
        \quad&+\frac{1}{\sigma^2(t)}\sum_{i=2}^Kw_i(\mb x)(\mb I-\mb U_i\tilde{\mb\Lambda}_{\sigma(t),i}\mb U_i^T)(\mb\mu_1-\mb\mu_i).
\end{align}

Note that:
\begin{itemize}[leftmargin=*]
    \item Guidance~\eqref{mixture cpc} resembles the CPC guidance $g_{cpc}(t)$ defined in~\eqref{all}. Different from the linear setting—where the CPC guidance contrasts the posterior covariance of the target class with a single unconditional posterior covariance, here it contrasts the posterior covariance of the target class with a softmax-weighted average of the posterior covariances of all classes.
    \item Guidance~\eqref{mixture mean-shift} resembles the mean-shift guidance $g_{mean}(t)$ defined in~\eqref{all}. Different from the linear setting where the mean-shift guidance approximately aligns with $\mb\mu_c-\mb\mu_{uc}$, the difference between the conditional and unconditional means, here it instead approximately aligns with a softmax-weighted average of the pairwise differences between the conditional mean (mean of the target class) and the means of every other class.
\end{itemize}

\section{Computing Resources}
\label{sec: computing resources}
All experiments are performed on A100 GPUs with 80 GB memory.

\end{document}